\documentclass[letterpaper, 10 pt, conference]{ieeeconf}  

\IEEEoverridecommandlockouts                              

\usepackage{graphicx}
\usepackage{amssymb}
\usepackage{amsfonts}
\usepackage{amsmath}
\usepackage{booktabs}
\usepackage{algorithm}
\usepackage{algorithmicx}
\usepackage[noend]{algpseudocode}
\usepackage{wrapfig}
\usepackage{bm}
\usepackage{bbm}
\usepackage{breqn}
\usepackage[font=small]{caption}
\usepackage[table]{xcolor}
\usepackage{array,multirow}
\usepackage{cprotect}
\usepackage{csquotes}
\usepackage{hyperref}

\usepackage{cite}

\algrenewcommand\algorithmicindent{0.5em}%

\usepackage{etoolbox}
\usepackage{tikz}
\usetikzlibrary{tikzmark}
\usetikzlibrary{calc}

\errorcontextlines\maxdimen

\newcommand{\ALGtikzmarkcolor}{black}
\newcommand{\ALGtikzmarkextraindent}{2pt}
\newcommand{\ALGtikzmarkverticaloffsetstart}{-.5ex}
\newcommand{\ALGtikzmarkverticaloffsetend}{-.5ex}
\makeatletter
\newcounter{ALG@tikzmark@tempcnta}

\newcommand\ALG@tikzmark@start{%
    \global\let\ALG@tikzmark@last\ALG@tikzmark@starttext%
    \expandafter\edef\csname ALG@tikzmark@\theALG@nested\endcsname{\theALG@tikzmark@tempcnta}%
    \tikzmark{ALG@tikzmark@start@\csname ALG@tikzmark@\theALG@nested\endcsname}%
    \addtocounter{ALG@tikzmark@tempcnta}{1}%
}

\def\ALG@tikzmark@starttext{start}
\newcommand\ALG@tikzmark@end{%
    \ifx\ALG@tikzmark@last\ALG@tikzmark@starttext
    \else
        \tikzmark{ALG@tikzmark@end@\csname ALG@tikzmark@\theALG@nested\endcsname}%
        \tikz[overlay,remember picture] \draw[\ALGtikzmarkcolor] let \p{S}=($(pic cs:ALG@tikzmark@start@\csname ALG@tikzmark@\theALG@nested\endcsname)+(\ALGtikzmarkextraindent,\ALGtikzmarkverticaloffsetstart)$), \p{E}=($(pic cs:ALG@tikzmark@end@\csname ALG@tikzmark@\theALG@nested\endcsname)+(\ALGtikzmarkextraindent,\ALGtikzmarkverticaloffsetend)$) in (\x{S},\y{S})--(\x{S},\y{E});%
    \fi
    \gdef\ALG@tikzmark@last{end}%
}

\apptocmd{\ALG@beginblock}{\ALG@tikzmark@start}{}{\errmessage{failed to patch}}
\pretocmd{\ALG@endblock}{\ALG@tikzmark@end}{}{\errmessage{failed to patch}}
\makeatother

\usepackage{todonotes}

\usepackage{epigraph}
\usepackage{mathtools}

\DeclarePairedDelimiterX{\kldivx}[2]{(}{)}{%
  #1\;\delimsize\|\;#2%
}
\newcommand{\kldiv}{D_{\mathrm{KL}}\kldivx}

\newcommand{\spec}{\tau^{\mathrm{spec}}}
\newcommand{\bs}{\mathbf{s}}
\newcommand{\bx}{\mathbf{x}}
\newcommand{\ba}{\mathbf{a}}

\newcommand{\bz}{\mathbf{z}}
\newcommand{\task}{\mathcal{T}}

\title{\LARGE \bf
ASHA: Assistive Teleoperation\\via Human-in-the-Loop Reinforcement Learning
}

\author{Sean Chen$^{1*}$, Jensen Gao$^{1*}$, Siddharth Reddy$^1$, Glen Berseth$^{1,2,3}$, Anca D. Dragan$^1$, Sergey Levine$^1$
\thanks{$^*$Equal Contribution, $^1$University of California, Berkeley, $^2$Universit\'e de Montr\'eal, $^3$MILA. Contact: \texttt{sgr@berkeley.edu}. Code, data, and videos available at \url{https://sites.google.com/view/asha-assist}.}%
        }

\begin{document}

\maketitle
\thispagestyle{empty}
\pagestyle{empty}

\begin{abstract}
Building assistive interfaces for controlling robots through arbitrary, high-dimensional, noisy inputs (e.g., webcam images of eye gaze) can be challenging, especially when it involves inferring the user's desired action in the absence of a natural `default' interface.
Reinforcement learning from online user feedback on the system's performance presents a natural solution to this problem, and enables the interface to adapt to individual users.
However, this approach tends to require a large amount of human-in-the-loop training data, especially when feedback is sparse.
We propose a hierarchical solution that learns efficiently from sparse user feedback: we use offline pre-training to acquire a latent embedding space of useful, high-level robot behaviors, which, in turn, enables the system to focus on using online user feedback to learn a mapping from user inputs to desired high-level behaviors.
The key insight is that access to a pre-trained policy enables the system to learn more from sparse rewards than a na\"ive RL algorithm: using the pre-trained policy, the system can make use of successful task executions to relabel, in hindsight, what the user actually meant to do during unsuccessful executions.
We evaluate our method primarily through a user study with 12 participants who perform tasks in three simulated robotic manipulation domains using a webcam and their eye gaze: flipping light switches, opening a shelf door to reach objects inside, and rotating a valve.
The results show that our method successfully learns to map 128-dimensional gaze features to 7-dimensional joint torques from sparse rewards in under 10 minutes of online training, and seamlessly helps users who employ different gaze strategies, while adapting to distributional shift in webcam inputs, tasks, and environments.
\end{abstract}

\section{Introduction} \label{intro}

\begin{figure}[t]
    \centering
    \includegraphics[width=\linewidth]{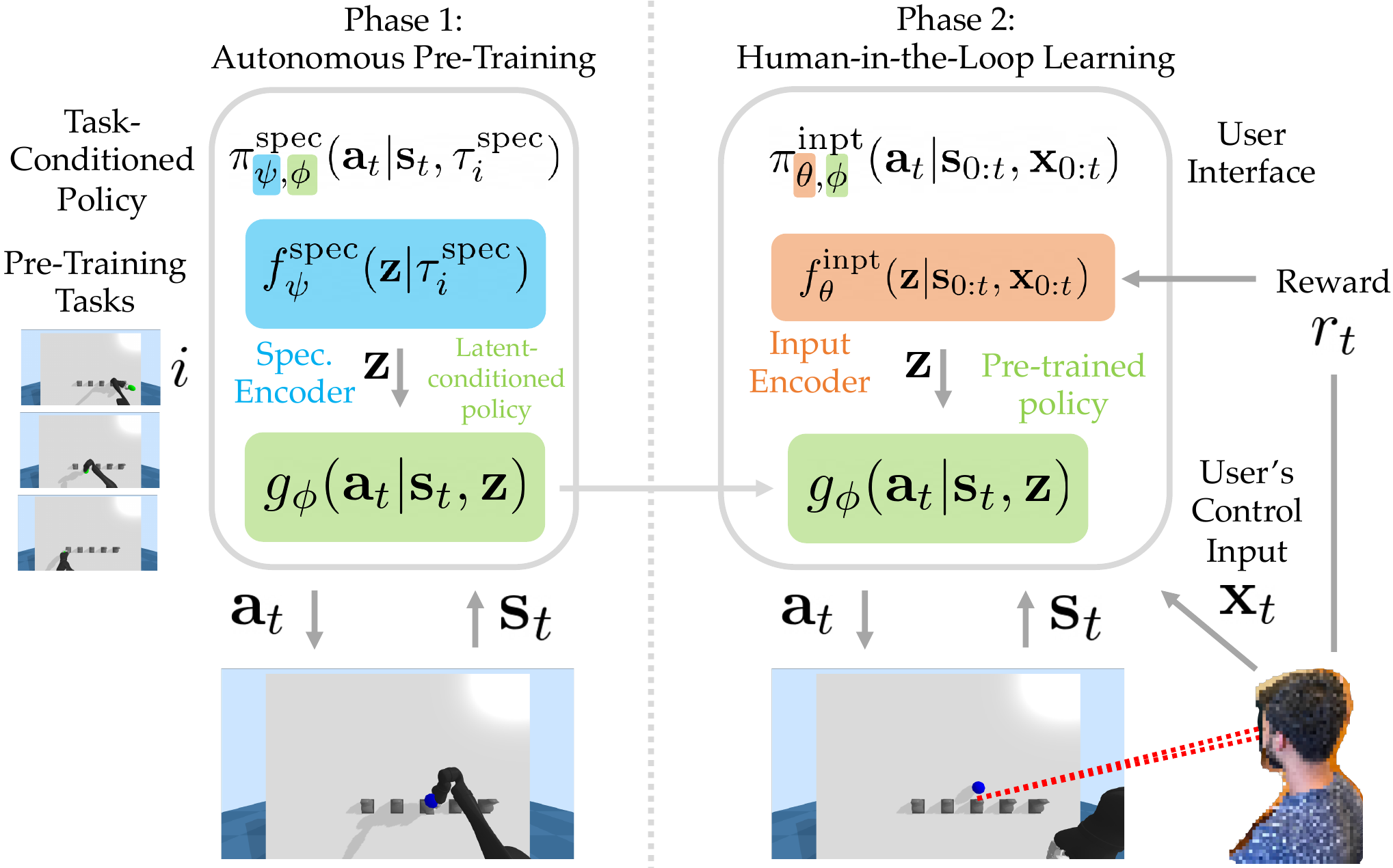}
    \caption{In this example, the user directs their gaze to control a wheelchair-mounted Jaco arm to push different light switches. During the autonomous pre-training phase, ASHA learns a task-conditioned policy $\pi_{\psi,\phi}^{\mathrm{spec}}$ that can flip various switches. This policy is decomposed into a latent variable model with two components: a specification encoder $f_{\psi}^{\mathrm{spec}}$, which maps a task specification $\spec$ (e.g., goal state) to a latent embedding $\bz$; and a latent-conditioned policy $g_{\phi}$. In phase 1, we jointly pre-train $f_{\psi}^{\mathrm{spec}}$ and $g_{\phi}$ to flip switches. In phase 2, we use human-in-the-loop RL to train an interface $\pi_{\theta,\phi}^{\mathrm{inpt}}$ that enables the user to control the arm using their eye gaze, and perform new tasks sampled from the same distribution as the pre-training tasks. To speed up learning, this interface is also represented as a latent variable model with two components: an input encoder $f_{\theta}^{\mathrm{inpt}}$, which maps the user's control input $\bx$ (e.g., webcam image) to a high-level, latent action $\bz$; and the pre-trained latent-conditioned policy $g_{\phi}$.}
    \label{fig:schematic}
\end{figure}

Shared-control teleoperation interfaces can help users control systems like robotic arms and wheelchairs more effectively~\cite{kim2006continuous,mcmullen2013demonstration,carlson2012collaborative,argall2016modular,javdani2017acting}.
For example, they can help users perform dexterous robotic manipulation tasks by automatically maximizing contact area with grasped objects~\cite{zhuang2019shared}, or enable control via complex user input streams like eye gaze~\cite{bien2004integration,aronson2018eye} and brain-computer interfaces~\cite{muelling2015autonomy}.
In this paper, we focus on the problem of efficiently training an interface to infer the user's desired action (e.g., robot arm motion) from an arbitrary, high-dimensional, noisy control input (e.g., webcam image of eye gaze).
This stands in contrast to prior work on shared autonomy that assumes the user already has a viable interface for direct teleoperation and only aims to improve the user's performance by minimally intervening in the user's actions to avoid collisions~\cite{broad2017learning,reddy2018shared,schaff2020residual,du2020ave}, preserving the reachability of potential goals~\cite{jeon2020shared}, or inferring goals and acting to reach them~\cite{hauser2013recognition,javdani2015shared,perez2015fast,koppula2016anticipating,muelling2017autonomy}.
Other prior methods do not require a direct teleop interface, and instead perform supervised calibration on paired examples of inputs and actions~\cite{gilja2012high,dangi2013design,dangi2014continuous,merel2015neuroprosthetic,wang2016learning,anumanchipalli2019speech,karamcheti2020learning,gaddy2020digital}.
However, this approach can also be limiting, in that it does not learn from the user's online interactions with the system during deployment, and as a result, does not improve with use or adapt to distributional shift in the user's inputs, tasks, and environments.

In this paper, we consider a different problem setting than the aforementioned prior work: instead of requiring a direct teleop interface or limiting data collection to explicit calibration phases, we elicit user feedback on the system's online performance and train the interface through reinforcement learning (RL) \cite{sutton2018reinforcement}.
Our adaptive interface observes the user's input, takes an action, receives a sparse, binary reward signal from the user at the end of each episode that indicates task success or failure, and learns to optimize this feedback.
This approach is appealing because it scales with regular use: the more the user uses the interface to perform the activities of daily living~\cite{ray2008people,mast2012user,petrich2021assistive}, the more competent and personalized the interface becomes.
Note that, in contrast to prior work on human-in-the-loop RL like COACH~\cite{macglashan2017interactive,arumugam2019deep}, TAMER~\cite{knox2009interactively,warnell2017deep}, and preference learning~\cite{dorsa2017active,christiano2017deep}, we aim to train an interface that enables the user to control the robot at test time and perform different tasks, instead of training the robot to autonomously perform a single task.
The main challenge is that, due to the sparsity of rewards, it can require a large amount of training data, which may be impractical for an individual user operating a physical robot.

We propose a hierarchical solution to this challenge: use offline pre-training to learn to perform potentially-useful tasks, then use online user feedback to learn a mapping from user inputs to robot behaviors (see Figure \ref{fig:schematic}).
In the pre-training phase, we train a task-conditioned policy to perform a wide variety of tasks without the user in the loop (e.g., rotating a valve to various target angles), and automatically discover useful, high-level robot behaviors in the process (e.g., rotating the valve clockwise or counter-clockwise).
Then, in the online learning phase, we bring the user into the loop, and use RL with sparse, user-provided rewards to learn how to interpret the user's inputs as desired high-level behaviors.
We leverage the pre-trained policy to extract more information from the user's sparse online rewards than standard RL algorithms: (a) when the user successfully completes a task, we observe information (e.g., the final state) that enables us to compute an optimal policy for that task in hindsight using our pre-trained task-conditioned policy, then train the interface to imitate that optimal policy; and (b) assuming that when the user fails, they attempt the same task again until they succeed, we can also relabel actions from failed trajectories with an optimal policy calculated in hindsight after an eventual success.
We call this algorithm \emph{ASsistive teleoperation via HumAn-in-the-loop reinforcement learning} (ASHA).

We primarily evaluate ASHA through a user study with 12 participants who use a webcam and their eye gaze to perform tasks in three simulated manipulation domains: flipping light switches, opening a shelf door to reach objects inside, and rotating a valve (see Figure \ref{fig:env-screenshots} for screenshots).
The results show that our method successfully learns to map 128-dimensional gaze features to 7-dimensional joint torques from sparse rewards in under 10 minutes of online training, while adapting to distributional shift in the user's webcam input caused by changes in ambient lighting and head position (Section \ref{drift-exp}); changes in the user's set of desired tasks, like the addition of a new light switch (Section \ref{goal-shift-exp}); and changes in environmental conditions, like whether a shelf door is initially open or closed (Section \ref{env-shift-exp}).
In both domains, ASHA increased success rates for the majority of users, compared to a non-adaptive baseline interface.
Even though users employed a variety of strategies to operate the interface --  e.g., looking directly at the target, looking at distant parts of the screen to indicate different targets, exaggerating their gaze to correct the robot, and dynamically guiding the robot to subgoals -- ASHA was able to seamlessly adapt to the different communication styles by learning from individual user data.

\section{Training an Assistive User Interface} \label{methods}

In our setting, the user cannot directly operate the robot.
Instead, the user relies on an assistive interface that infers the user's intended action from available inputs, such as webcam images of eye gaze, or signals recorded by a brain implant.
We do not require prior knowledge of how to parse the user's input, and instead treat the user's input as a raw, undifferentiated bitstream.
The user's desired task is typically not directly observable to the robot, and this desired task may change between episodes.
As such, we formulate the assistance problem as a partially-observable Markov decision process (POMDP) \cite{kaelbling1998planning}.
The state consists of the state of the environment $\bs_t$ (e.g., the position and orientation of the robot) and the user's desired task $\task$ (e.g., flipping a particular light switch). 
The observation consists of the state of the environment $\bs_t$ and the user's control input $\bx_t$ (e.g., an image of their eyes that captures gaze direction), but not the task $\task$.
The user's control input $\bx_t$ communicates their intent to the robot.
We do not assume access to the user's desired task $\task$, since this can be difficult for the user to specify.
Instead, we elicit a sparse, binary reward signal $r_t \in \{0, 1\}$ from the user, in the form of a button press that indicates task success or failure at the end of each episode.
We aim to learn an interface $\pi^{\mathrm{inpt}}(\ba_t|\bs_{0:t},\bx_{0:t})$ that optimizes this user-provided reward feedback.
We also aim to minimize the number of human interactions required to learn this interface.

Our approach to this problem is outlined in Figure \ref{fig:schematic}.
Training an assistive interface through human-in-the-loop RL with sparse rewards typically requires many hours of interactions with users, in part due to the difficulty of simultaneously learning to control the environment and infer the user's intent~\cite{reddy2018shared}.
However, in typical teleoperation tasks, there are aspects of controlling the environment that can be learned separately from the user.
Hence, we decompose the problem into two phases: (1) pre-training a policy $g(\ba_t|\bs_t,\bz)$ that is parameterized by a high-level latent variable $\bz$ and can perform potentially-useful tasks; and (2) learning a mapping $f^{\mathrm{inpt}}(\bz|\bs_{0:t},\bx_{0:t})$ from the user's control inputs $\bx$ to the user's desired high-level behavior $\bz$.

\subsection{Phase 1: Autonomous Pre-Training of a Task-Conditioned Policy} \label{pretraining}

In many teleoperation domains, we can conservatively define a task distribution that covers a wide variety of behaviors that the user may potentially want to execute in the future -- e.g., opening and closing cupboards in a kitchen, or flipping light switches on a wall -- and then pre-train the robot to perform those tasks, without the user in the loop.
Our final system is not necessarily limited to selecting from among these pre-training tasks.
Rather, the space of skills acquired in phase 1 is meant to act as a kind of `basis' for the tasks that the user might want to perform in phase 2, and continuous input from the user will be used to infer the desired behavior in terms of this basis.
This space can also be viewed as a reparameterization of the policy space: instead of searching over all possible policy parameters during the human-in-the-loop learning phase, the system will search over high-level behaviors in the latent space acquired during this autonomous pre-training phase.

During phase 1, we assume the ability to sample tasks $\task_i \sim p(\task)$, a specification $\spec_i$ of each task, and a reward function $R_i(\bs_t, \ba_t)$ for each task.
These reward functions $R_i$ are only used for pre-training, and are not required during phase 2 of human-in-the-loop learning, when the user will perform new, unknown tasks drawn from the same distribution $p(\task)$.
Also note that the specification $\spec_i$ does not have to be a full trajectory, but merely a representation of the task that can be extracted from a successful trajectory -- in our experiments, we define a set of goal-reaching tasks, set each specification $\spec_i$ to be the 3D position of the target object or 1D target angle of the valve, and define a reward function $R_i$ for each task.
To ensure that we learn a basis of skills, rather than a separate policy for each of the pre-training tasks, we follow prior work~\cite{hausman2018learning} and represent the robot's policy as a latent variable model,
\begin{equation} \label{eqn:pretrain-decomp}
\pi_{\psi,\phi}^{\mathrm{spec}}(\ba_t|\bs_t;\spec_i) \triangleq \mathbb{E}_{\bz \sim f_{\psi}^{\mathrm{spec}}(\bz|\spec_i)}[g_{\phi}(\ba_t|\bs_t,\bz)],
\end{equation}
where $f_{\psi}^{\mathrm{spec}}$ is the `specification encoder', $g_{\phi}$ is the `latent-conditioned policy', $\pi_{\psi,\phi}^{\mathrm{spec}}$ is the composition of $f_{\psi}^{\mathrm{spec}}$ and $g_{\phi}$, $\bz \in \mathbb{R}^d$ is a latent variable that characterizes the task (we set $d = 3$ in our experiments), the prior distribution of $\bz$ is the standard normal distribution $\mathcal{N}(\mathbf{0}, I_d)$, and the action $\ba_t$ is conditionally independent of the specification $\spec_i$ given the state $\bs_t$ and latent embedding $\bz$.
At the beginning of each pre-training episode, we sample a task $\task_i \sim p(\task)$.
We then jointly pre-train the latent-conditioned policy $g_{\phi}$ and specification encoder $f_{\psi}^{\mathrm{spec}}$ to optimize the task rewards $R_i$ using RL -- in our implementation, we use the soft actor-critic algorithm (SAC) \cite{haarnoja2018soft}.
An important consequence of the latent variable model in Equation \ref{eqn:pretrain-decomp} is that, in addition to optimizing the task rewards $R_i$, we regularize the latent embedding $\bz$ to its prior distribution $\mathcal{N}(\mathbf{0}, I_d)$ using a variational information bottleneck (VIB) \cite{alemi2016deep,achille2018information}.
The VIB encourages the model to learn a smooth, compressed latent space that shares information across tasks, and encourages the specification encoder $f_{\psi}^{\mathrm{spec}}$ to discard task-irrelevant information about the specification $\spec_i$ from the embedding $\bz$ -- this is critical to phase 2 of our method, because it prevents the interface from attempting to infer these task-irrelevant details from the user's control inputs (see \textbf{Q2} in Section \ref{ab-exp}).

\subsection{Phase 2: Human-in-the-Loop Reinforcement Learning of a User Interface} \label{hitl-phase}

Now that we have acquired a latent embedding space of high-level robot behaviors (the left half of Figure \ref{fig:schematic}), we turn to the problem of learning an interface that maps user inputs to desired high-level behaviors (the right half of Figure \ref{fig:schematic}).
We represent the interface as a latent variable model that reuses the pre-trained latent-conditioned policy $g_{\phi}$,
\begin{equation} \label{eqn:online-decomp}
\pi_{\theta,\phi}^{\mathrm{inpt}}(\ba_t|\bs_{0:t},\bx_{0:t}) \triangleq \mathbb{E}_{\bz \sim f_{\theta}^{\mathrm{inpt}}(\bz|\bs_{0:t},\bx_{0:t})}[g_{\phi}(\ba_t|\bs_t,\bz)],
\end{equation}
where $f_{\theta}^{\mathrm{inpt}}$ is the `input encoder', $\pi_{\theta,\phi}^{\mathrm{inpt}}$ is the composition of $f_{\theta}^{\mathrm{inpt}}$ and $g_{\phi}$, and the prior distribution of $\bz$ is the standard normal distribution $\mathcal{N}(\mathbf{0},I_d)$.
Note that the input encoder $f_{\theta}^{\mathrm{inpt}}$ differs from the specification encoder $f_{\psi}^{\mathrm{spec}}$ learned in the pre-training phase: $f_{\theta}^{\mathrm{inpt}}$ reads in the user's control input $\bx$ (e.g., gaze), while $f_{\psi}^{\mathrm{spec}}$ takes a specification $\spec$ (e.g., goal state) as input instead.
Since the user's inputs $\bx$ are only partial observations of the state variable $\mathcal{T}$ that defines the task, the interface $\pi_{\theta,\phi}^{\mathrm{inpt}}$ is generally conditioned on the full sequence of states $\bs_{0:t}$ and inputs $\bx_{0:t}$.
However, in our switch and bottle experiments, we find that conditioning on only the most recent state $\bs_t$ and input $\bx_t$ works well empirically.

Given the latent variable model in Equation \ref{eqn:online-decomp}, we train $f_{\theta}^{\mathrm{inpt}}$ through RL from user feedback.
Recent work in this area~\cite{gao2021xt} suggests a straightforward method: assign a reward of 1 to successes, 0 to failures, and run a standard RL algorithm that essentially imitates the successful trajectories while down-weighting the failed trajectories.
Unfortunately, due to the sparsity of the rewards, this approach would typically require a prohibitive amount of human interaction (see \textbf{Q3} in Section \ref{ab-exp}).
However, we contribute a novel insight that makes the method practical: we can extract more information from successful trajectories, by not simply imitating the actions that were actually taken (since some of them may be suboptimal), but instead imitating an optimal policy for the task that was completed.
We can also extract more information from failed trajectories in the same manner, if we assume that when the user fails to perform a task, they reset the robot to its initial state (e.g., retract the robotic arm on their wheelchair back to its mount), and try to perform the same task again and again until they succeed.
We now operationalize these two ideas, then arrive at our final method.

\noindent\textbf{Learning efficiently from successes in hindsight.}
Instead of simply imitating a successful trajectory, we imitate an optimal policy conditioned on task information extracted from the successful trajectory.
Let $\mathcal{D}$ denote the set of successful trajectories.
From each of these successes, we extract a specification $\spec$ of the user's desired task at the time -- e.g., in the switch and bottle domains, we set $\spec$ to be the final 3D position of the object manipulated in the successful trajectory, analogous to the pre-training phase in Section \ref{pretraining}.
We then combine this task specification $\spec$ with the pre-trained task-conditioned policy $\pi_{\psi,\phi}^{\mathrm{spec}}$ to represent the optimal policy $\pi_{\psi,\phi}^{\mathrm{spec}}(\ba_t|\bs_t,\spec)$ via Equation \ref{eqn:pretrain-decomp}.
The key idea is to match the interface $\pi_{\theta,\phi}^{\mathrm{inpt}}$ with the optimal policy $\pi_{\psi,\phi}^{\mathrm{spec}}$, by optimizing the loss,
\begin{align} \label{eqn:match-pol}
\mathcal{L}(\theta) = \sum_{\tau \in \mathcal{D}, t} & \kldiv{\pi_{\psi,\phi}^{\mathrm{spec}}(\cdot|\bs_t,\spec)}{\pi_{\theta,\phi}^{\mathrm{inpt}}(\cdot|\bs_{0:t},\bx_{0:t})} \nonumber \\
 + \beta &  \kldiv{f_{\theta}^{\mathrm{inpt}}(\cdot|\bs_{0:t},\bx_{0:t})}{\mathcal{N}(\mathbf{0}, I_d)},
\end{align}
where the second term is the VIB for the latent variable model in Equation \ref{eqn:online-decomp}, and $\beta$ is a regularization constant.
By minimizing the divergence between the policies induced by the input encoder $f_{\theta}^{\mathrm{inpt}}$ and the pre-trained specification encoder $f_{\psi}^{\mathrm{spec}}$, we force $f_{\theta}^{\mathrm{inpt}}$ to infer a latent embedding that induces the same low-level action distribution as the embedding inferred by $f_{\psi}^{\mathrm{spec}}$.
This helps to reduce the amount of human interaction required to train the system (see \textbf{Q5} in Section \ref{ab-exp}).
Note that optimizing Equation \ref{eqn:match-pol} does not necessarily force both encoders to produce the same embedding, since different embeddings can induce the same low-level action distribution.
This keeps our method flexible, and enables the user to guide the robot to subgoals.

\noindent\textbf{Learning efficiently from failures in hindsight.}
Instead of simply treating failed trajectories as examples of behavior that achieved zero reward, we take the final, successful trajectory at the end of a string of failed trajectories that attempted to perform the same task, extract a task specification $\spec$ from this successful trajectory, compute the optimal policy $\pi_{\psi,\phi}^{\mathrm{spec}}(\ba_t|\bs_t,\spec)$ for all the states in the success \emph{and the failures}, and optimize the loss in Equation \ref{eqn:match-pol}.
The key idea is that successful episodes enable us to compute the optimal policy for the most recent failure episodes, because a string of failures and eventual success are all attempts to perform the same task.
This helps to minimize the human interaction required to train the system (see \textbf{Q4} in Section \ref{ab-exp}).

\subsection{Algorithm Summary} \label{alg-summ}

\begin{algorithm}[t]
\begin{algorithmic}[1]
\small
\State{$g_{\phi}, f_{\psi}^{\mathrm{spec}} \leftarrow \mathrm{RL}(\{\spec_i, R_i\}_i)$ \Comment{pre-train the latent-conditioned policy and specification encoder autonomously}}
\While{true}
  \State{$\task \sim p(\task)$ \Comment{user chooses a task}}
  \State{$\mathcal{D} \leftarrow []$ \Comment{initialize empty list of trajectories for current task}}
  \While{robot has not succeeded at task $\task$ yet}
  \State{$\tau \leftarrow []$ \Comment{initialize empty trajectory}}
  \State{$\bs_0 \sim p(\bs_0)$ \Comment{reset environment}}
  \For{$t \in \{0, 1, 2, ..., T - 1\}$}
    \State{$\bx_t \leftarrow \text{user's control input}$}
    \State{$\ba_t \sim \pi_{\theta,\phi}^{\mathrm{inpt}}(\ba_t|\bs_{0:t},\bx_{0:t})$ \Comment{robot performs action}}
    \State{$\tau$.append($\bs_t, \bx_t$)}
    \State{$\bs_{t+1} \sim p(\bs_{t+1} | \bs_t, \ba_t)$ \Comment{environment evolves}}
  \EndFor
  \State{$\mathcal{D}$.append($\tau$) \Comment{store trajectory (even if a failure)}}
  \EndWhile
  \State{$\spec \leftarrow$ final, successful trajectory $\tau$ in $\mathcal{D}$}
  \State{$\theta \leftarrow \theta - \nabla_{\theta} \sum_{\tau \in \mathcal{D}, t} \kldiv{\pi_{\psi,\phi}^{\mathrm{spec}}(\cdot|\bs_t,\spec)}{\pi_{\theta,\phi}^{\mathrm{inpt}}(\cdot|\bs_{0:t},\bx_{0:t})}$\\~~~~~~~$+ \mathrm{VIB}(\theta)$ \Comment{update input encoder}} \label{lst:line:theta-update}
\EndWhile
\end{algorithmic}
\caption{Assistive Teleoperation via Human-in-the-Loop Reinforcement Learning (ASHA)}
\label{alg:ASHA-alg}
\end{algorithm}

Our complete assistive teleoperation method is summarized in Algorithm \ref{alg:ASHA-alg}.
We initially pre-train the latent-conditioned policy $g_{\phi}$ and specification encoder $f_{\psi}^{\mathrm{spec}}$ with a set of task specifications and reward functions $\{\spec_i, R_i\}_i$ using a standard RL algorithm with a VIB -- our implementation constructs several goal-reaching tasks and pre-trains on them with SAC.
We then begin training the input encoder $f_{\theta}^{\mathrm{inpt}}$ with the user in the loop.
First, the user decides on a task $\task$, which we assume is sampled from the same distribution $p(\task)$ as the pre-training tasks.
At each timestep $t$, the environment generates the next state $\bs_t$, and the user provides the system with input $\bx_t$.
After seeing the input $\bx_t$, the robot takes an action $\ba_t$ sampled from the interface $\pi_{\theta,\phi}^{\mathrm{inpt}}$ defined by the input encoder $f_{\theta}^{\mathrm{inpt}}$ and the pre-trained latent-conditioned policy $g_{\phi}$ via Equation \ref{eqn:online-decomp}.
At the end of each trajectory, we ask the user whether the robot succeeded or failed.
If the robot fails, we reset and assume the user attempts to perform the same task again.
If the robot succeeds, we take the successful trajectory, extract a task specification $\spec$ from it, use the pre-trained specification encoder $f_{\psi}^{\mathrm{spec}}$ and latent-conditioned policy $g_{\phi}$ to define an optimal policy for the task via Equation \ref{eqn:pretrain-decomp}, and train the input encoder $f_{\theta}^{\mathrm{inpt}}$ to induce actions that match that optimal policy, by optimizing the loss in Equation \ref{eqn:match-pol}.
We find that training $f_{\theta}^{\mathrm{inpt}}$ to convergence in line \ref{lst:line:theta-update} using mini-batch stochastic gradient descent on all past data, including data $\mathcal{D}$ from previous tasks $\mathcal{T}$, works well empirically.
The user then decides on a new task, and we repeat with the updated input encoder $f_{\theta}^{\mathrm{inpt}}$.

\section{User Studies} \label{user-studies}

\begin{figure}[t]
    \centering
    \includegraphics[width=\linewidth]{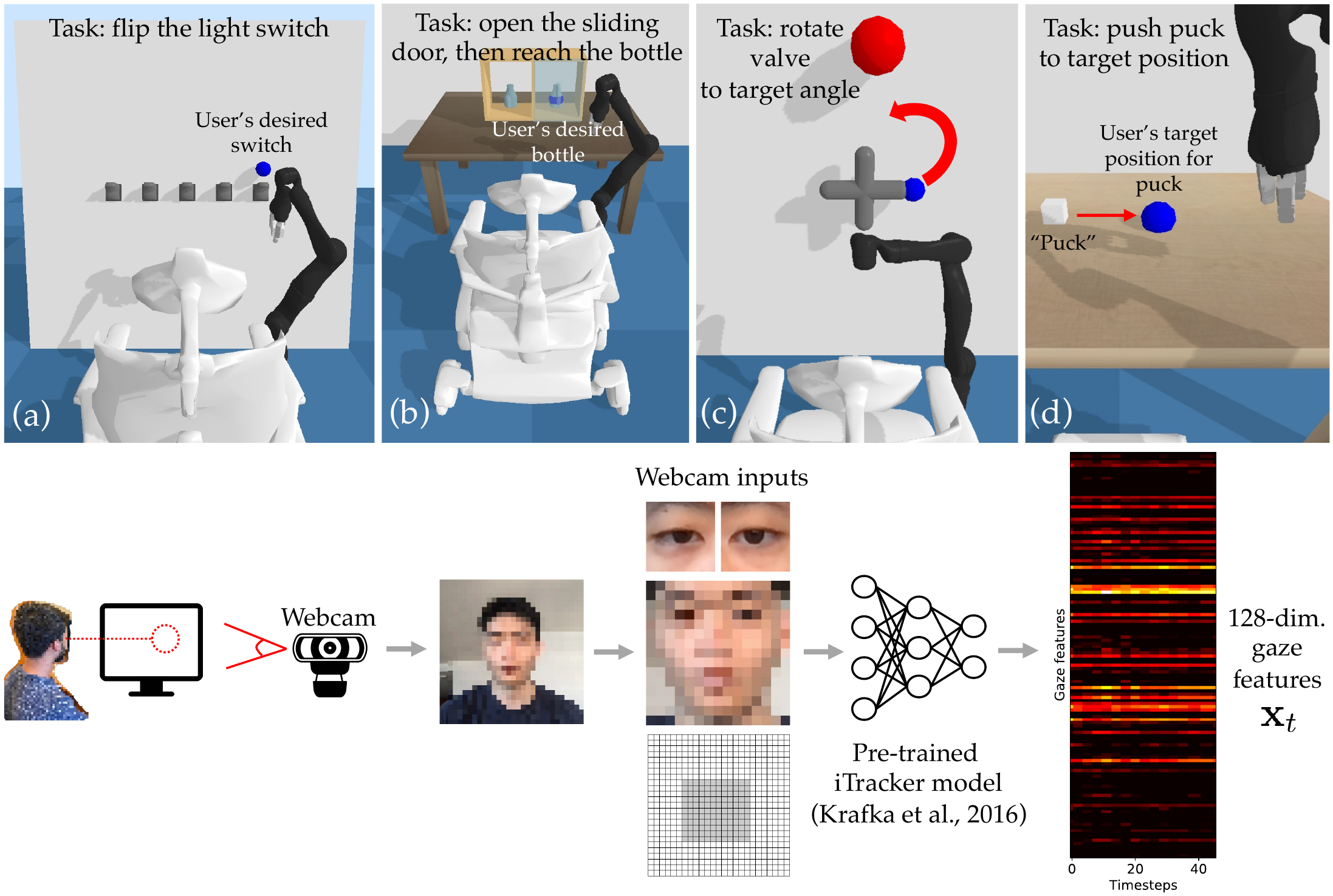}
    \caption{The user sees the simulated environment through the point of view of someone sitting in the wheelchair, and directs their gaze to communicate what they want the robot to do. In the switch domain (a), the user must push down on the blue lever. In the bottle domain (b), the user must open the sliding door if necessary, then reach for the blue bottle. In the valve domain (c), the user must rotate the valve so that the blue tip points at the red sphere. In the puck domain (d), the user must push the white puck to the blue target.}
    \label{fig:env-screenshots}
\end{figure}

In our experiments, we evaluate to what extent ASHA can adapt to the user's inputs (Section \ref{drift-exp}), to users that want to perform new tasks (Section \ref{goal-shift-exp}), and to changes in the environment (Section \ref{env-shift-exp}).
We conduct a user study with 12 participants who control a simulated 7-DoF Jaco robotic arm using gaze (see Figure \ref{fig:env-screenshots}).
The interface receives 128-dimensional feature vectors that represent the user's webcam image inputs $\bx_t$, and outputs 7-dimensional joint torques as actions $\ba_t$.
The users perform tasks in three simulated manipulation domains implemented with the PyBullet real-time physics simulator~\cite{coumans2016pybullet} using assets from Assistive Gym~\cite{erickson2020assistive}: flipping light switches, opening a shelf to reach objects inside, and rotating a valve.

\subsection{Adapting to Distributional Shift in Gaze Inputs} \label{drift-exp}

In this experiment, we aim to test ASHA's ability to improve over time by learning from user feedback.
We compare to a non-adaptive baseline interface that is initially calibrated via supervised learning, but does not adapt during deployment (analogous to the prior work discussed in Section \ref{intro}).
To train this baseline interface, we collect paired data by showing a small number of pre-recorded videos of the robot autonomously performing tasks to the user, and recording the user's passive gaze inputs as they watch the videos.
We show 2 videos per task in each domain, totalling 6 videos in switch, 8 in bottle, and 8 in valve.
We then train the baseline's input encoder $f_{\theta_0}^{\mathrm{inpt}}$ on the objective in Equation \ref{eqn:match-pol}, treating the paired data as a set of successful trajectories $\mathcal{D}$. 
We refer to this supervised learning procedure as `calibration'.
Note that this implicitly assumes that the user's passive inputs are equivalent to their active control inputs, which is often not the case in practice~\cite{cunningham2011closed,willett2019principled}.
To improve the initial performance and sample efficiency of our method, we initialize ASHA's input encoder $f_{\theta}^{\mathrm{inpt}}$ with the calibrated baseline parameters $\theta_0$, and initialize ASHA's replay buffer with the same paired data that was used to calibrate the baseline.
We measure the online performance of both methods by asking the user to complete particular tasks (e.g., flipping the switch indicated in blue), and computing the success rate of the user's first attempt at each task (including subsequent attempts would introduce selection effects for difficult tasks).
We calibrate and evaluate on the same distribution of tasks: in the switch domain, a uniform distribution over flipping one of the three switches in the middle; and in the bottle domain, a uniform distribution over reaching one of the two bottles.
To establish a lower bound on performance, we also compare to a baseline that randomly samples a latent $\bz$ and executes the policy $g_{\phi}(\ba_t|\bs_t,\bz)$, without taking any user input.

\begin{figure}[t]
    \centering
    \includegraphics[width=0.75\linewidth]{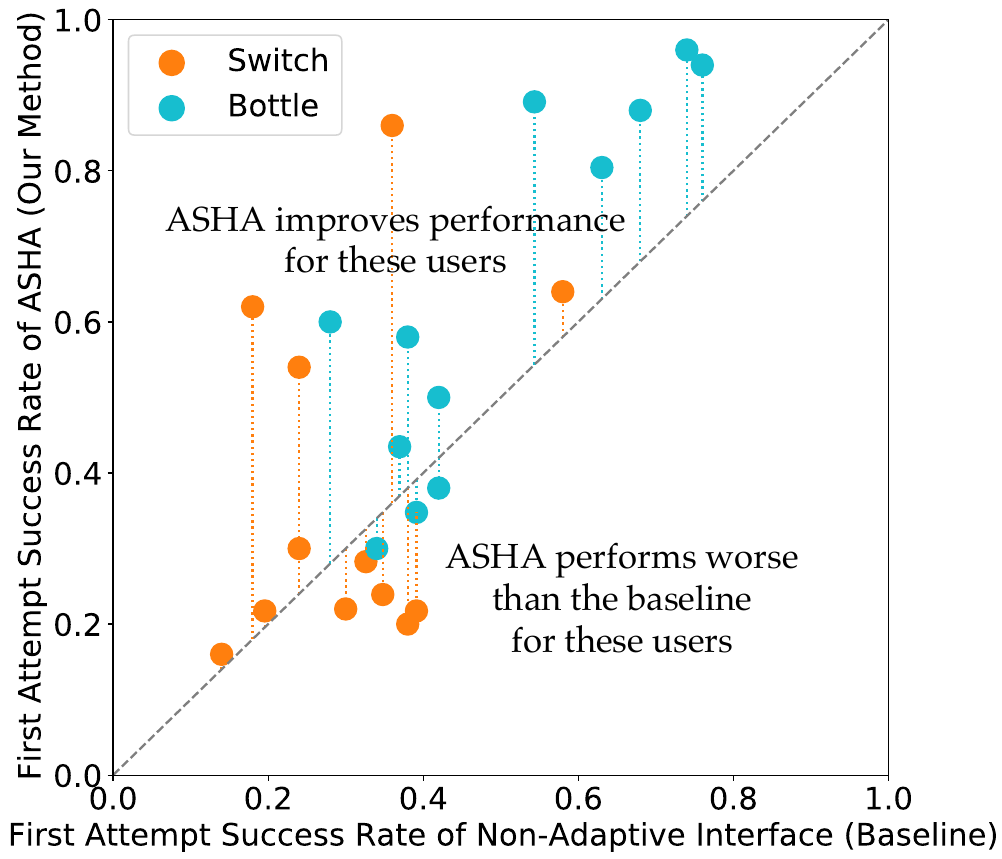}
    \caption{Each circle represents the success rate for one participant, averaged over 50 online episodes (11 minutes).}
    \label{fig:user-study-main}
  \vspace{-0.15in}
\end{figure}

The results in Figure \ref{fig:user-study-main} show that ASHA improves the success rates of the majority of users, relative to the non-adaptive baseline.
ASHA initially performs the same as the non-adaptive baseline, executing coherent but undesirable behaviors like moving toward the wrong target, then begins to outperform the baseline after 20 online episodes of RL.
One potential explanation for the gap between ASHA and the non-adaptive baseline is that most users have a substantial distribution mismatch between passive and active inputs, and that ASHA helps those users by fine-tuning on active inputs instead of only initially calibrating on passive inputs.
Another possibility is that ASHA adapts to changes in ambient lighting or head position over time, while the non-adaptive baseline performs increasingly worse over time.
We ran a one-way repeated measures ANOVA on the success rates from the baseline and ASHA conditions, with the presence of ASHA as a factor, and found that $f(1, 11) = 8.26, p < .05$ in the switch domain, and $f(1, 11) = 7.28, p < .05$ in the bottle domain.
Subjective evaluations corroborate these results: users reported feeling more in control of the robot with ASHA compared to the baseline.

\subsection{Learning to Perform the User's Desired Tasks} \label{goal-shift-exp}

The previous experiment showed that ASHA can adapt to distributional shift in the user's gaze input.
In this next experiment, we show that ASHA can also adapt to individual differences in the user's desired task distribution.
In the switch domain in particular,  we calibrate the input encoder on paired data generated from one distribution of tasks -- a uniform distribution over the 2nd and 3rd switches from the left -- then evaluate online on a different distribution of tasks -- a uniform distribution over the 2nd, 3rd, and 4th switches from the left.
This is challenging, since examples of the 4th switch being pressed are not included in the calibration data.
RL offers a natural solution to this problem by fine-tuning the model on the user's online attempts to perform new tasks.
The results in Figure \ref{fig:user-study-shift}a show that ASHA can indeed adapt to the new task distribution, substantially improving upon its initial success rate by the end of the online training period.

\subsection{Adapting to a Changing Environment} \label{env-shift-exp}

\begin{figure}
  \begin{center}
  \caption{Shift in Task Distribution or Environment}
    \includegraphics[width=\linewidth]{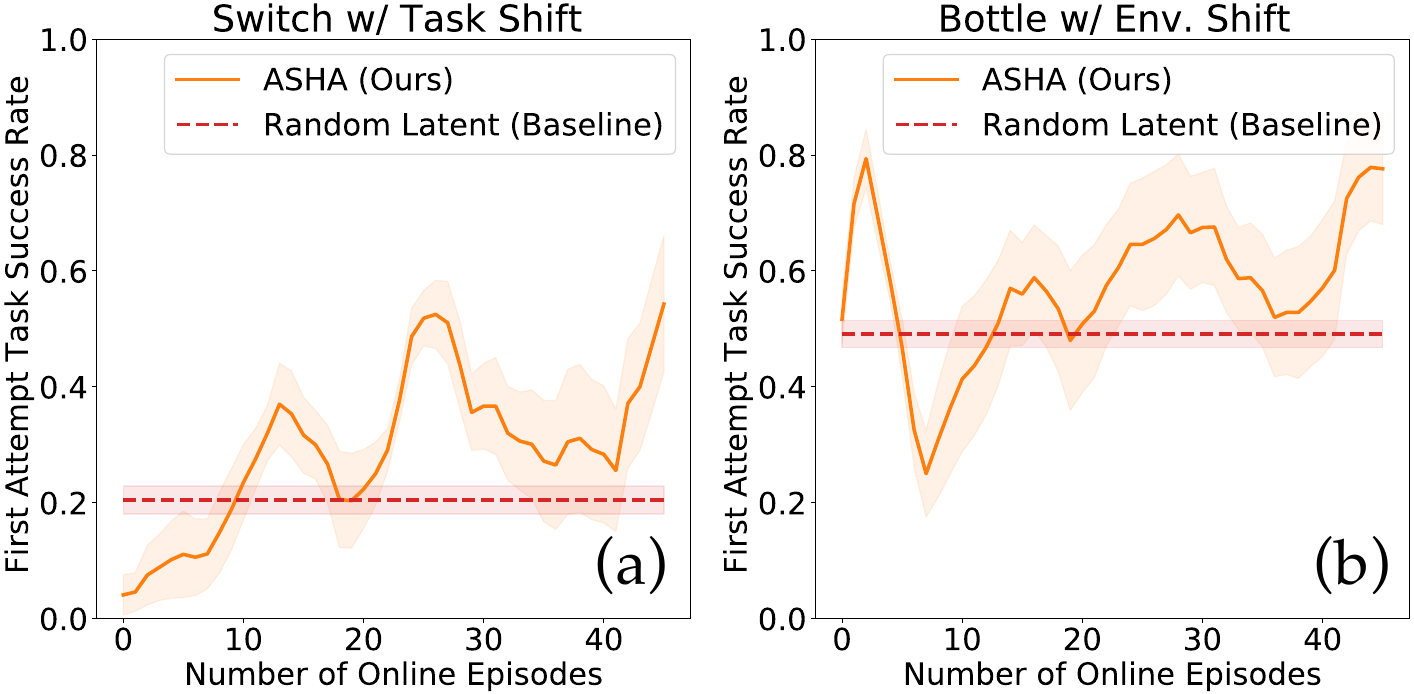}
    \label{fig:user-study-shift}
  \end{center}
  \vspace{-0.15in}
\end{figure}
Adaptation is useful, not only because the user's input might drift or their desired tasks might be novel relative to the training tasks, but also because the environment may have changed since the interface was previously calibrated.
RL again offers a natural solution to this problem by incorporating new experiences into its replay memory as the user interacts with their changing environment.
To illustrate this idea, we run an experiment in the bottle domain in which we calibrate on paired data where the sliding door never covers the desired bottle, then evaluate online in scenarios where the sliding door may randomly cover the desired bottle.
Figure \ref{fig:user-study-shift}b shows that ASHA adapts to the new environmental conditions, increasing the success rate over time.

\section{Simulation Experiments} \label{sim-exp}

\subsection{Ablation Study} \label{ab-exp}

\begin{table}
  \begin{center}
    \caption{Ablation Experiments}
   \begin{tabular}{lll}
    \toprule
     & Switch & Bottle \\
    \midrule
    Random Latent (Baseline) & $ 0.19 \pm 0.02 $ & $ 0.44 \pm 0.02 $ \\
    Non-Adaptive (Baseline) & $ 0.50 \pm 0.05 $ & $ 0.53 \pm 0.02 $  \\
    \hline
    \textbf{ASHA (Ours)} & $\mathbf{ 0.83 \pm 0.02 }$ & $\mathbf{ 0.79 \pm 0.03 }$ \\
    ASHA w/ Det. Input Enc. (\textbf{Q1}) & $ 0.70 \pm 0.03 $ & $ 0.73 \pm 0.02 $ \\
    ASHA w/ Det. Pre-train Enc. (\textbf{Q2}) & $ 0.66 \pm 0.06 $ & $ 0.46 \pm 0.03 $ \\
    SAC from Scratch (\textbf{Q3}) & $ 0.00 \pm 0.00 $ & $ 0.00 \pm 0.00 $ \\
    ASHA w/o Failure Relabeling (\textbf{Q4}) & $ 0.54 \pm 0.03 $ & $ 0.55 \pm 0.02 $ \\
    ASHA w/ Latent Regression (\textbf{Q5}) & $ 0.41 \pm 0.04 $ & $ 0.57 \pm 0.02 $ \\
    \bottomrule
  \end{tabular}
  \vspace{5pt}
  \caption*{Success rates across 100 episodes and 10 random seeds}
   \label{tab:ablation-means}
  \end{center}
\end{table}
To run ablation experiments at a scale that would be impractical in a user study, we simulate user input $\bx \in \mathbb{R}^3$ as the 3D position of the target switch or bottle with i.i.d. isotropic Gaussian noise added at each timestep.
We seek to answer the following questions.
\textbf{Q1}: Does sampling from a stochastic input encoder $f_{\theta}^{\mathrm{inpt}}$ improve exploration, relative to a deterministic encoder?
\textbf{Q2}: Does pre-training with a VIB improve downstream performance during human-in-the-loop learning, relative to pre-training without a VIB?
\textbf{Q3}: Does pre-training the latent-conditioned policy $g_{\phi}$ speed up human-in-the-loop learning, relative to end-to-end training the interface from scratch online?
\textbf{Q4}: Does relabeling failures speed up human-in-the-loop learning, relative to ignoring failures and only training on successes?
\textbf{Q5}: Does regressing onto the optimal policy in Equation \ref{eqn:match-pol} perform better than regressing onto sampled latents that led to a success?
The results in Table \ref{tab:ablation-means} show that all the ablated variants of ASHA perform worse than the full ASHA method, suggesting that sampling from a stochastic input encoder $f_{\theta}^{\mathrm{inpt}}$ improves exploration (\textbf{Q1}), pre-training with a VIB and reusing the pre-trained latent-conditioned policy $g_{\phi}$ speed up downstream learning (\textbf{Q2}, \textbf{Q3}), relabeling failures makes human-in-the-loop learning more efficient (\textbf{Q4}), and regressing onto an optimal policy is more effective than regressing onto sampled latents (\textbf{Q5}).

\subsection{Demonstration on Continuous Task Spaces}

The switch and bottle environments tested in the previous experiments have discrete task spaces: the user either wants to flip one of five switches, or reach one of two bottles.
However, ASHA can also effectively assist users who have continuous task spaces.
To demonstrate this capability, we ran experiments with simulated users \textbf{and three expert human users} who rotate a valve to a desired target angle $\theta \sim \mathrm{Unif}(0, 2\pi)$ -- i.e., a continuous, 1D task space (see Figure \ref{fig:env-screenshots}c).
In addition to the expert human input, we tested various types of simulated user inputs, including static inputs that noisily encode the target angle, dynamic inputs that noisily encode subgoals on a path to the goal state, and directional inputs that noisily indicate whether to rotate clockwise, counter-clockwise, or remain.
The results in Figure \ref{fig:valve-puck-results}a show that ASHA learns to perform the desired task with an 80\% success rate when an expert human provides input (pink), a 90\% success rate when we simulate user input that is static and simple to decode (orange), and performs substantially better than a random-latent baseline policy (red) when we simulate user input that encodes nearby subgoals (blue) or desired direction of rotation (gray).

\subsection{Demonstration on Structured User Inputs}

\begin{figure}[t]
  \begin{center}
  \caption{Continuous Task Spaces and Structured User Inputs}
    \includegraphics[width=\linewidth]{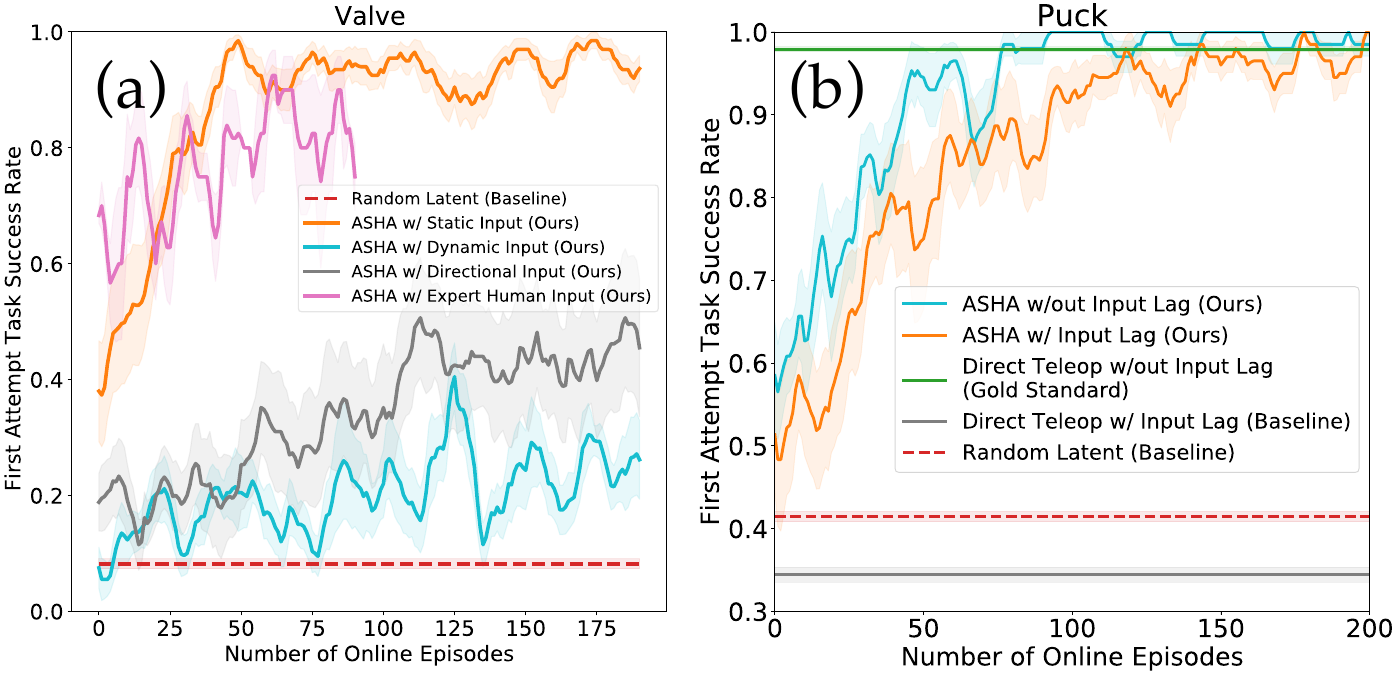}
    \label{fig:valve-puck-results}
  \end{center}
\end{figure}

This paper focuses on the problem of interpreting raw user inputs like webcam images as commands.
However, ASHA can also be used to assist users who already have access to a direct teleoperation interface for translating raw user inputs into robot actions, but still require help to perform their desired tasks.
To illustrate this capability, we ran experiments with simulated users who push a puck on a table to a desired target position, which is sampled uniformly at random from a continuous, 2D task space (see Figure \ref{fig:env-screenshots}d).
We simulated user inputs by training an oracle policy to perform the pushing task, then adding lag to the oracle actions, which models real-world conditions like network latency -- as a result, the user inputs to ASHA are (suboptimal) 7-dimensional joint torques.
The results in Figure \ref{fig:valve-puck-results}b show that, when the user's input is laggy, ASHA (orange) achieves substantially higher success rates than the direct teleop interface alone (gray), eventually reaching the performance of a hypothetical direct teleop interface that receives user input without lag (green).

\section{Discussion}

We presented a system that efficiently trains an adaptive interface through RL from sparse user feedback.
Our user studies in three simulated robotic manipulation domains show that, in under 10 minutes of online learning, our method can adapt to distributional shift in webcam inputs, tasks, and environments.
One limitation of our method is that it assumes the ability to sample pre-training tasks and accompanying reward functions (see Section \ref{pretraining}).
Future work could use a self-supervised RL algorithm to discover a latent skill space without a pre-determined distribution of tasks~\cite{eysenbach2018diversity,nair2020contextual,lynch2020learning}.
Despite this limitation, ASHA illustrates how RL can provide a general mechanism for efficiently adapting user interfaces to individual needs; not only for assistive robotic teleoperation, but also potentially for other domains, such as brain-computer interfaces for speech decoding~\cite{guenther2009wireless,bocquelet2016real}.

\section{Acknowledgements}

Thanks to members of the InterACT and RAIL labs at UC Berkeley for feedback on this project.
This work was supported by NVIDIA Graduate Fellowship, Berkeley Existential Risk Initiative, AFOSR FA9550-17-1-0308, Weill Neurohub, NSF CAREER, CIFAR, Office of Naval Research, Army Research Office, ARL DCIST CRA W911NF-17-2-0181, the National Science Foundation under IIS-1651843, and GPU computing resources from NVIDIA.

\bibliographystyle{unsrt}
\bibliography{main}

\begin{thebibliography}{10}

\bibitem{kim2006continuous}
Hyun~K Kim, J~Biggs, W~Schloerb, M~Carmena, Mikhail~A Lebedev, Miguel~AL
  Nicolelis, and Mandayam~A Srinivasan.
\newblock Continuous shared control for stabilizing reaching and grasping with
  brain-machine interfaces.
\newblock {\em IEEE Transactions on Biomedical Engineering}, 2006.

\bibitem{mcmullen2013demonstration}
David~P McMullen, Guy Hotson, Kapil~D Katyal, Brock~A Wester, Matthew~S Fifer,
  Timothy~G McGee, Andrew Harris, Matthew~S Johannes, R~Jacob Vogelstein,
  Alan~D Ravitz, et~al.
\newblock Demonstration of a semi-autonomous hybrid brain--machine interface
  using human intracranial {EEG}, eye tracking, and computer vision to control
  a robotic upper limb prosthetic.
\newblock {\em IEEE Transactions on Neural Systems and Rehabilitation
  Engineering}, 2013.

\bibitem{carlson2012collaborative}
Tom Carlson and Yiannis Demiris.
\newblock Collaborative control for a robotic wheelchair: evaluation of
  performance, attention, and workload.
\newblock {\em IEEE Transactions on Systems, Man, and Cybernetics}, 2012.

\bibitem{argall2016modular}
Brenna~D Argall.
\newblock Modular and adaptive wheelchair automation.
\newblock In {\em Experimental Robotics}, 2016.

\bibitem{javdani2017acting}
Shervin Javdani.
\newblock {\em Acting under Uncertainty for Information Gathering and Shared
  Autonomy}.
\newblock PhD thesis, Carnegie Mellon University, 2017.

\bibitem{zhuang2019shared}
Katie~Z Zhuang, Nicolas Sommer, Vincent Mendez, Saurav Aryan, Emanuele
  Formento, Edoardo D’Anna, Fiorenzo Artoni, Francesco Petrini, Giuseppe
  Granata, Giovanni Cannaviello, et~al.
\newblock Shared human--robot proportional control of a dexterous myoelectric
  prosthesis.
\newblock {\em Nature Machine Intelligence}, 2019.

\bibitem{bien2004integration}
Zeungnam Bien, Myung-Jin Chung, Pyung-Hun Chang, Dong-Soo Kwon, Dae-Jin Kim,
  Jeong-Su Han, Jae-Hean Kim, Do-Hyung Kim, Hyung-Soon Park, Sang-Hoon Kang,
  et~al.
\newblock Integration of a rehabilitation robotic system ({KARES II}) with
  human-friendly man-machine interaction units.
\newblock {\em Autonomous Robots}, 2004.

\bibitem{aronson2018eye}
Reuben~M Aronson, Thiago Santini, Thomas~C K{\"u}bler, Enkelejda Kasneci,
  Siddhartha Srinivasa, and Henny Admoni.
\newblock Eye-hand behavior in human-robot shared manipulation.
\newblock In {\em IEEE Conference on Human-Robot Interaction}, 2018.

\bibitem{muelling2015autonomy}
Katharina Muelling, Arun Venkatraman, Jean-Sebastien Valois, John Downey,
  Jeffrey Weiss, Shervin Javdani, Martial Hebert, Andrew~B Schwartz, Jennifer~L
  Collinger, and J~Andrew Bagnell.
\newblock Autonomy infused teleoperation with application to {BCI}
  manipulation.
\newblock {\em arXiv preprint arXiv:1503.05451}, 2015.

\bibitem{broad2017learning}
Alexander Broad, Todd~David Murphey, and Brenna~Dee Argall.
\newblock Learning models for shared control of human-machine systems with
  unknown dynamics.
\newblock In {\em Robotics: Science and Systems}, 2017.

\bibitem{reddy2018shared}
Siddharth Reddy, Anca~D Dragan, and Sergey Levine.
\newblock Shared autonomy via deep reinforcement learning.
\newblock {\em arXiv preprint arXiv:1802.01744}, 2018.

\bibitem{schaff2020residual}
Charles Schaff and Matthew~R Walter.
\newblock Residual policy learning for shared autonomy.
\newblock {\em arXiv preprint arXiv:2004.05097}, 2020.

\bibitem{du2020ave}
Yuqing Du, Stas Tiomkin, Emre Kiciman, Daniel Polani, Pieter Abbeel, and Anca
  Dragan.
\newblock {AvE}: Assistance via empowerment.
\newblock {\em arXiv preprint arXiv:2006.14796}, 2020.

\bibitem{jeon2020shared}
Hong~Jun Jeon, Dylan~P Losey, and Dorsa Sadigh.
\newblock Shared autonomy with learned latent actions.
\newblock {\em arXiv preprint arXiv:2005.03210}, 2020.

\bibitem{hauser2013recognition}
Kris Hauser.
\newblock Recognition, prediction, and planning for assisted teleoperation of
  freeform tasks.
\newblock {\em Autonomous Robots}, 2013.

\bibitem{javdani2015shared}
Shervin Javdani, Siddhartha~S Srinivasa, and J~Andrew Bagnell.
\newblock Shared autonomy via hindsight optimization.
\newblock {\em arXiv preprint arXiv:1503.07619}, 2015.

\bibitem{perez2015fast}
Claudia P{\'e}rez-D'Arpino and Julie~A Shah.
\newblock Fast target prediction of human reaching motion for cooperative
  human-robot manipulation tasks using time series classification.
\newblock In {\em IEEE International Conference on Robotics and Automation},
  2015.

\bibitem{koppula2016anticipating}
Hema~S Koppula and Ashutosh Saxena.
\newblock Anticipating human activities using object affordances for reactive
  robotic response.
\newblock {\em IEEE Transactions on Pattern Analysis and Machine Intelligence},
  2016.

\bibitem{muelling2017autonomy}
Katharina Muelling, Arun Venkatraman, Jean-Sebastien Valois, John~E Downey,
  Jeffrey Weiss, Shervin Javdani, Martial Hebert, Andrew~B Schwartz, Jennifer~L
  Collinger, and J~Andrew Bagnell.
\newblock Autonomy infused teleoperation with application to brain computer
  interface controlled manipulation.
\newblock {\em Autonomous Robots}, 2017.

\bibitem{gilja2012high}
Vikash Gilja, Paul Nuyujukian, Cindy~A Chestek, John~P Cunningham, M~Yu Byron,
  Joline~M Fan, Mark~M Churchland, Matthew~T Kaufman, Jonathan~C Kao, Stephen~I
  Ryu, et~al.
\newblock A high-performance neural prosthesis enabled by control algorithm
  design.
\newblock {\em Nature neuroscience}, 2012.

\bibitem{dangi2013design}
Siddharth Dangi, Amy~L Orsborn, Helene~G Moorman, and Jose~M Carmena.
\newblock Design and analysis of closed-loop decoder adaptation algorithms for
  brain-machine interfaces.
\newblock {\em Neural computation}, 2013.

\bibitem{dangi2014continuous}
Siddharth Dangi, Suraj Gowda, Helene~G Moorman, Amy~L Orsborn, Kelvin So,
  Maryam Shanechi, and Jose~M Carmena.
\newblock Continuous closed-loop decoder adaptation with a recursive maximum
  likelihood algorithm allows for rapid performance acquisition in
  brain-machine interfaces.
\newblock {\em Neural computation}, 2014.

\bibitem{merel2015neuroprosthetic}
Josh Merel, David Carlson, Liam Paninski, and John~P Cunningham.
\newblock Neuroprosthetic decoder training as imitation learning.
\newblock {\em arXiv preprint arXiv:1511.04156}, 2015.

\bibitem{wang2016learning}
Sida~I Wang, Percy Liang, and Christopher~D Manning.
\newblock Learning language games through interaction.
\newblock {\em arXiv preprint arXiv:1606.02447}, 2016.

\bibitem{anumanchipalli2019speech}
Gopala~K Anumanchipalli, Josh Chartier, and Edward~F Chang.
\newblock Speech synthesis from neural decoding of spoken sentences.
\newblock {\em Nature}, 2019.

\bibitem{karamcheti2020learning}
Siddharth Karamcheti, Dorsa Sadigh, and Percy Liang.
\newblock Learning adaptive language interfaces through decomposition.
\newblock {\em arXiv preprint arXiv:2010.05190}, 2020.

\bibitem{gaddy2020digital}
David Gaddy and Dan Klein.
\newblock Digital voicing of silent speech.
\newblock {\em arXiv preprint arXiv:2010.02960}, 2020.

\bibitem{sutton2018reinforcement}
Richard~S Sutton and Andrew~G Barto.
\newblock {\em Reinforcement learning: An introduction}.
\newblock 2018.

\bibitem{ray2008people}
C{\'e}line Ray, Francesco Mondada, and Roland Siegwart.
\newblock What do people expect from robots?
\newblock In {\em IEEE/RSJ International Conference on Intelligent Robots and
  Systems}, 2008.

\bibitem{mast2012user}
Marcus Mast, Michael Burmester, Katja Kr{\"u}ger, Sascha Fatikow, Georg
  Arbeiter, Birgit Graf, Gernot Kronreif, Lucia Pigini, David Facal, and Renxi
  Qiu.
\newblock User-centered design of a dynamic-autonomy remote interaction concept
  for manipulation-capable robots to assist elderly people in the home.
\newblock {\em Journal of Human-Robot Interaction}, 2012.

\bibitem{petrich2021assistive}
Laura Petrich, Jun Jin, Masood Dehghan, and Martin Jagersand.
\newblock Assistive arm and hand manipulation: How does current research
  intersect with actual healthcare needs?
\newblock {\em arXiv preprint arXiv:2101.02750}, 2021.

\bibitem{macglashan2017interactive}
James MacGlashan, Mark~K Ho, Robert Loftin, Bei Peng, David Roberts, Matthew~E
  Taylor, and Michael~L Littman.
\newblock Interactive learning from policy-dependent human feedback.
\newblock {\em arXiv preprint arXiv:1701.06049}, 2017.

\bibitem{arumugam2019deep}
Dilip Arumugam, Jun~Ki Lee, Sophie Saskin, and Michael~L Littman.
\newblock Deep reinforcement learning from policy-dependent human feedback.
\newblock {\em arXiv preprint arXiv:1902.04257}, 2019.

\bibitem{knox2009interactively}
W~Bradley Knox and Peter Stone.
\newblock Interactively shaping agents via human reinforcement: The {TAMER}
  framework.
\newblock In {\em International Conference on Knowledge Capture}, 2009.

\bibitem{warnell2017deep}
Garrett Warnell, Nicholas Waytowich, Vernon Lawhern, and Peter Stone.
\newblock Deep {TAMER}: Interactive agent shaping in high-dimensional state
  spaces.
\newblock {\em arXiv preprint arXiv:1709.10163}, 2017.

\bibitem{dorsa2017active}
Dorsa Sadigh, Anca~D Dragan, Shankar Sastry, and Sanjit~A Seshia.
\newblock Active preference-based learning of reward functions.
\newblock In {\em Robotics: Science and Systems}, 2017.

\bibitem{christiano2017deep}
Paul~F Christiano, Jan Leike, Tom Brown, Miljan Martic, Shane Legg, and Dario
  Amodei.
\newblock Deep reinforcement learning from human preferences.
\newblock In {\em Neural Information Processing Systems}, 2017.

\bibitem{kaelbling1998planning}
Leslie~Pack Kaelbling, Michael~L Littman, and Anthony~R Cassandra.
\newblock Planning and acting in partially observable stochastic domains.
\newblock {\em Artificial Intelligence}, 1998.

\bibitem{hausman2018learning}
Karol Hausman, Jost~Tobias Springenberg, Ziyu Wang, Nicolas Heess, and Martin
  Riedmiller.
\newblock Learning an embedding space for transferable robot skills.
\newblock In {\em International Conference on Learning Representations}, 2018.

\bibitem{haarnoja2018soft}
Tuomas Haarnoja, Aurick Zhou, Pieter Abbeel, and Sergey Levine.
\newblock Soft actor-critic: Off-policy maximum entropy deep reinforcement
  learning with a stochastic actor.
\newblock In {\em International Conference on Machine Learning}, 2018.

\bibitem{alemi2016deep}
Alexander~A Alemi, Ian Fischer, Joshua~V Dillon, and Kevin Murphy.
\newblock Deep variational information bottleneck.
\newblock {\em arXiv preprint arXiv:1612.00410}, 2016.

\bibitem{achille2018information}
Alessandro Achille and Stefano Soatto.
\newblock Information dropout: Learning optimal representations through noisy
  computation.
\newblock {\em IEEE Transactions on Pattern Analysis and Machine Intelligence},
  2018.

\bibitem{gao2021xt}
Jensen Gao, Siddharth Reddy, Glen Berseth, Nicholas Hardy, Nikhilesh Natraj,
  Karunesh Ganguly, Anca Dragan, and Sergey Levine.
\newblock {X2T}: Training an x-to-text typing interface with online learning
  from user feedback.
\newblock In {\em International Conference on Learning Representations}, 2021.

\bibitem{coumans2016pybullet}
Erwin Coumans and Yunfei Bai.
\newblock Py{B}ullet, a {P}ython module for physics simulation for games,
  robotics and machine learning.
\newblock 2016.

\bibitem{erickson2020assistive}
Zackory Erickson, Vamsee Gangaram, Ariel Kapusta, C~Karen Liu, and Charles~C
  Kemp.
\newblock Assistive gym: A physics simulation framework for assistive robotics.
\newblock In {\em IEEE International Conference on Robotics and Automation},
  2020.

\bibitem{cunningham2011closed}
John~P Cunningham, Paul Nuyujukian, Vikash Gilja, Cindy~A Chestek, Stephen~I
  Ryu, and Krishna~V Shenoy.
\newblock A closed-loop human simulator for investigating the role of feedback
  control in brain-machine interfaces.
\newblock {\em Journal of Neurophysiology}, 2011.

\bibitem{willett2019principled}
Francis~R Willett, Daniel~R Young, Brian~A Murphy, William~D Memberg,
  Christine~H Blabe, Chethan Pandarinath, Sergey~D Stavisky, Paymon Rezaii, Jad
  Saab, Benjamin~L Walter, et~al.
\newblock Principled {BCI} decoder design and parameter selection using a
  feedback control model.
\newblock {\em Scientific Reports}, 2019.

\bibitem{eysenbach2018diversity}
Benjamin Eysenbach, Abhishek Gupta, Julian Ibarz, and Sergey Levine.
\newblock Diversity is all you need: Learning skills without a reward function.
\newblock {\em arXiv preprint arXiv:1802.06070}, 2018.

\bibitem{nair2020contextual}
Ashvin Nair, Shikhar Bahl, Alexander Khazatsky, Vitchyr Pong, Glen Berseth, and
  Sergey Levine.
\newblock Contextual imagined goals for self-supervised robotic learning.
\newblock In {\em Conference on Robot Learning}, 2020.

\bibitem{lynch2020learning}
Corey Lynch, Mohi Khansari, Ted Xiao, Vikash Kumar, Jonathan Tompson, Sergey
  Levine, and Pierre Sermanet.
\newblock Learning latent plans from play.
\newblock In {\em Conference on Robot Learning}, 2020.

\bibitem{guenther2009wireless}
Frank~H Guenther, Jonathan~S Brumberg, E~Joseph Wright, Alfonso Nieto-Castanon,
  Jason~A Tourville, Mikhail Panko, Robert Law, Steven~A Siebert, Jess~L
  Bartels, Dinal~S Andreasen, et~al.
\newblock A wireless brain-machine interface for real-time speech synthesis.
\newblock {\em PloS one}, 2009.

\bibitem{bocquelet2016real}
Florent Bocquelet, Thomas Hueber, Laurent Girin, Christophe Savariaux, and
  Blaise Yvert.
\newblock Real-time control of an articulatory-based speech synthesizer for
  brain computer interfaces.
\newblock {\em PLoS Computational Biology}, 2016.

\bibitem{mcgovern1998hierarchical}
Amy McGovern, Doina Precup, Balaraman Ravindran, Satinder Singh, and Richard~S
  Sutton.
\newblock Hierarchical optimal control of {MDP}s.
\newblock 1998.

\bibitem{parr1998reinforcement}
Ronald Parr and Stuart~J Russell.
\newblock Reinforcement learning with hierarchies of machines.
\newblock In {\em Neural Information Processing Systems}, 1998.

\bibitem{HAC}
Andrew Levy, Robert Platt, and Kate Saenko.
\newblock Hierarchical actor-critic.
\newblock {\em arXiv preprint arXiv:1712.00948}, 2017.

\bibitem{HIRO}
Ofir Nachum, Shixiang~Shane Gu, Honglak Lee, and Sergey Levine.
\newblock Data-efficient hierarchical reinforcement learning.
\newblock In {\em Neural Information Processing Systems}, 2018.

\bibitem{HIPPO}
Alexander Li, Carlos Florensa, Ignasi Clavera, and Pieter Abbeel.
\newblock Sub-policy adaptation for hierarchical reinforcement learning.
\newblock In {\em International Conference on Learning Representations}, 2020.

\bibitem{nachum2018near}
Ofir Nachum, Shixiang Gu, Honglak Lee, and Sergey Levine.
\newblock Near-optimal representation learning for hierarchical reinforcement
  learning.
\newblock {\em arXiv preprint arXiv:1810.01257}, 2018.

\bibitem{sensinger2009adaptive}
Jonathon~W Sensinger, Blair~A Lock, and Todd~A Kuiken.
\newblock Adaptive pattern recognition of myoelectric signals: exploration of
  conceptual framework and practical algorithms.
\newblock {\em IEEE Transactions on Neural Systems and Rehabilitation
  Engineering}, 2009.

\bibitem{pilarski2011online}
Patrick~M Pilarski, Michael~R Dawson, Thomas Degris, Farbod Fahimi, Jason~P
  Carey, and Richard~S Sutton.
\newblock Online human training of a myoelectric prosthesis controller via
  actor-critic reinforcement learning.
\newblock In {\em IEEE International Conference on Rehabilitation Robotics},
  2011.

\bibitem{ross2011reduction}
St{\'e}phane Ross, Geoffrey Gordon, and Drew Bagnell.
\newblock A reduction of imitation learning and structured prediction to
  no-regret online learning.
\newblock In {\em International Conference on Artificial Intelligence and
  Statistics}, 2011.

\bibitem{krafka2016eye}
Kyle Krafka, Aditya Khosla, Petr Kellnhofer, Harini Kannan, Suchendra
  Bhandarkar, Wojciech Matusik, and Antonio Torralba.
\newblock Eye tracking for everyone.
\newblock In {\em IEEE Conference on Computer Vision and Pattern Recognition},
  2016.

\bibitem{kingma2014adam}
Diederik~P Kingma and Jimmy Ba.
\newblock Adam: A method for stochastic optimization.
\newblock {\em arXiv preprint arXiv:1412.6980}, 2014.

\bibitem{nair2020accelerating}
Ashvin Nair, Murtaza Dalal, Abhishek Gupta, and Sergey Levine.
\newblock Accelerating online reinforcement learning with offline datasets.
\newblock {\em arXiv preprint arXiv:2006.09359}, 2020.

\bibitem{rakelly2019efficient}
Kate Rakelly, Aurick Zhou, Chelsea Finn, Sergey Levine, and Deirdre Quillen.
\newblock Efficient off-policy meta-reinforcement learning via probabilistic
  context variables.
\newblock In {\em International Conference on Machine Learning}, 2019.

\end{thebibliography}

\clearpage

\appendix 

\subsection{Additional Related Work} \label{more-related-work}

ASHA's use of a pre-training phase to learn a low-level policy $g_{\phi}$ that accelerates downstream learning of a high-level user interface $f_{\theta}^{\mathrm{inpt}}$ resembles hierarchical RL methods~\cite{mcgovern1998hierarchical,parr1998reinforcement}, which generally aim to improve exploration and credit assignment by dividing the original Markov decision process (MDP) into simpler high- and low-level MDPs. 
Recent work in this area focuses on solving the divided MDPs concurrently~\cite{HAC,HIRO,HIPPO} and combining the results~\cite{nachum2018near}. 
Our problem setting and assumptions are markedly different, in that the engineered reward function used to pre-train the low-level policy in phase 1 of ASHA may differ substantially from the user-provided reward that is used to train the interface online in phase 2, while standard hierarchical methods assume that these two reward functions are equivalent.

Prior work on myoelectric interfaces for prosthetic limb control explores unsupervised learning~\cite{sensinger2009adaptive} and RL~\cite{pilarski2011online} methods for online adaptation to the user's EMG signals, but only evaluates on simple motion-tracking tasks with small, discrete state spaces, whereas our work focuses on more complex manipulation tasks with large, continuous state spaces (see Figure \ref{fig:env-screenshots} and Appendix \ref{representations}).

ASHA's approach to relabeling trajectories with an optimal policy (see line \ref{lst:line:theta-update} in Algorithm \ref{alg:ASHA-alg}) is analogous to the DAgger imitation learning algorithm, which relabels on-policy trajectories of an imitation policy with expert action labels~\cite{ross2011reduction}.

\subsection{Implementation Details} \label{imp-details}

\subsubsection{Timeouts} \label{timeout}

To ensure that the user studies do not get stalled on a single task, if the user does not succeed at a given task after 5 episodes of attempts, we `timeout' and ask them to move on to a new task.
We do not learn from these 5 episodes in our method, since we do not have a success that we can use to compute the optimal policy in hindsight (see Section \ref{hitl-phase}).
Across all methods tested and all experimental conditions in the user studies, timeouts occurred in 6\% of tasks in the bottle domain, and 20\% of tasks in the switch domain.

\subsubsection{Simulator Setup for Switch and Bottle Domains} \label{switch-bottle-setup}

We setup our switch and bottle domains using assets from Assistive Gym~\cite{erickson2020assistive}.
The sizes of the simulated Jaco arm and wheelchair are proportional to those of their real-world counterparts.
In the switch domain, the switches are placed 0.22 units of distance apart horizontally.
The horizontal position of the switches vary uniformly at random within an interval of 0.3 units, but all switches share the same positional noise (i.e., all switches move together).
The distance between the user and the wall varies uniformly at random within an interval of 0.2 units during the pre-training and calibration phases -- during online evaluation, it only varies within half that range. 
The initial arm end-effector position varies within a uniform box of size (1, 0.2, 0.2).
In the bottle domain, the bottles are placed 0.3 units of distance apart horizontally. 
The horizontal position of the table varies uniformly at random within an interval of 0.4 units.
The bottles always maintain the same relative position with respect to the table.
The initial arm end-effector position varies uniformly at random within a box of size (0.8, 0.2, 0.2).
Due to the noise added to the positions of switches and bottles at the beginning of each episode, the positions for different switches and bottles can overlap across episodes (e.g., switch 1 may be located very close to the previous position of switch 2 in a past episode).
While the arm is reset to its initial position at the start of each episode, regardless of the success or failure of the previous episode, the other conditions of the environment (e.g., the positions of the switches, table, or wall) are only reset after a successful episode or a timeout.  
The arm is reset by first sampling a 3D position, then solving for the joint positions of the arm using an inverse kinematics function.

The environments use a frame skip of 5 steps, and a maximum episode length of 200 steps (approximately 13 seconds of wall-clock time).

\subsubsection{Representation of States, Actions, Task Specifications, and Latent Embeddings} \label{representations}

In all domains, the low-level action $\ba \in \mathbb{R}^7$ consists of 7 joint forces for the Jaco arm. 
Each action dimension is clipped to (-0.25, 0.25) before being executed in the simulator.
In the switch domain, the state $\bs \in \mathbb{R}^{48}$ consists of the 7 joint positions of the arm, the 3D position and 4D orientation of the end effector, and the 3D position and 1D angle of each of the 5 switches.
In the bottle domain, the state $\bs \in \mathbb{R}^{37}$ consists of the 7 joint positions of the arm, the 3D position and 4D orientation of the end effector, the 3D position of each of the 2 bottles, and the 3D position of door handle.
We set each specification $\spec \in \mathbb{R}^3$ to be the 3D position of the target switch or bottle.
In both domains, we set the dimensionality of the latent embedding space to $d = 3$.

\subsubsection{Recording the User's Eye Gaze} \label{gaze-details}

With a standard webcam, we record a 224x224x3 segmented image of the user's face, 224x224x3 image of each of the user's eyes, and a 25x25 binary grid that characterizes the overall position of the face in the webcam image, then feed these as input to a pre-trained iTracker model~\cite{krafka2016eye} (see Figure \ref{fig:env-screenshots}c). 
We treat the 128-dimensional activations of the last linear layer in the pre-trained iTracker neural network as the user's control input $\bx \in \mathbb{R}^{128}$.
Gaze is recorded by having an asynchronous thread that takes the webcam image, segments it, feeds it into the iTracker network, then extracts the features, sets it equal to the most recent features, and terminates.
On every step of the environment simulator, we restart the gaze update thread if it has terminated, then pull the most recent gaze features and treat them as the user input $\bx_t$ for that timestep.

\subsubsection{Network Architecture and Optimization} \label{arch-and-opt}

In all phases (i.e., pre-training, calibration, and online learning), we use the Adam optimizer~\cite{kingma2014adam} with a batch size of 256 to train our models.
We use the same network architecture to represent all encoders $f_{\psi}^{\mathrm{spec}}$, $f_{\theta_0}^{\mathrm{inpt}}$, and $f_{\theta}^{\mathrm{inpt}}$: a feedforward network with ReLU activations and a single hidden layer of 64 units.

In all phases, we set the regularization constant $\beta$ for the VIB (e.g., see Equation \ref{eqn:match-pol}) to 0.01.

In the pre-training phase, we run SAC with default hyperparameters: a 2-layer, 256-unit feedforward network to represent both the policy and Q-function, a learning rate of $3 \cdot 10^{-4}$, a reward scale of 1, automatic entropy tuning with a heuristic of setting the target entropy to the negative dimensionality of the action space, a target Q-function update period of 1, Polyack update $\tau$ set to $5 \cdot 10^{-3}$, epochs of 1000 environment steps followed by 1000 training steps, 1000 steps before training starts, and a replay buffer size of $5 \cdot 10^5$ in the switch domain and $2 \cdot 10^7$ in the bottle domain.
We pre-train for 1000 epochs in the switch domain, and for 3150 epochs in the bottle domain.
In the bottle domain, we initialize the pre-training replay buffer with 5000 demonstrations obtained from a scripted agent.
When we execute the pre-training policy $\pi_{\psi,\phi}^{\mathrm{spec}}$, we feed the expected value of the specification encoder output $\bz$ to the latent-conditioned policy $g_{\phi}$, instead of sampling from the posterior of the encoder.

During the online learning phase, we set the learning rate to $5 \cdot 10^{-4}$, perform 1000 gradient updates on the calibration data (which is typically sufficient for convergence), keep the calibration data in the replay buffer during online learning, perform 100 gradient updates after each successful episode (which is typically sufficient for convergence), and do not limit the size of the replay buffer (i.e., never discard old data).

We use the same random seeds for each user, and use a different random seed for each method and experimental condition within a given user.
We use 10 different random seeds for each ablation experiment, and use the same 10 seeds across the ablations.

We set the optimal policy $\pi_{\psi,\phi}^{\mathrm{spec}}$ used in Equation \ref{eqn:match-pol} during the online learning phase to be deterministic.
Implicitly assuming that the interface $\pi_{\theta,\phi}^{\mathrm{inpt}}$ has some fixed, diagonal covariance $I\sigma^2$, this enables us to simplify the KL divergence loss between the two policies $\pi_{\psi,\phi}^{\mathrm{spec}}$ and $\pi_{\theta,\phi}^{\mathrm{inpt}}$ in Equation \ref{eqn:match-pol} to the mean-squared error loss between the mean actions outputted by both policies.

We do not use a recurrent input encoder $f_{\theta}^{\mathrm{inpt}}(\bz|\bs_{0:t},\bx_{0:t})$, and instead use a feedforward encoder $f_{\theta}^{\mathrm{inpt}}(\bz|\bs_t,\bx_t)$ that operates on only the most recent state $\bs_t$ and user input $\bx_t$.

To simplify the optimization of Equation \ref{eqn:match-pol}, we do not integrate over the posterior distribution of latents $\bz$ when computing the optimal policy $\pi_{\psi,\phi}^{\mathrm{spec}}$ and interface $\pi_{\theta,\phi}^{\mathrm{inpt}}$, and instead only feed the expected value of $\bz$ to the latent-conditioned policy $g_{\phi}$ -- i.e., we represent the optimal policy as $g_{\phi}(\ba_t|\bs_t,\mathbb{E}_{\bz \sim f_{\psi}^{\mathrm{spec}}(\bz|\spec)}[\bz])$ (instead of Equation \ref{eqn:pretrain-decomp}), and the interface as $g_{\phi}(\ba_t|\bs_t,\mathbb{E}_{\bz \sim f_{\theta}^{\mathrm{inpt}}(\bz|\bs_{0:t},\bx_{0:t})}[\bz])$ (instead of Equation \ref{eqn:online-decomp}).

\subsubsection{Pre-Training Tasks} \label{pretrain-rews}

In the switch domain, we have 5 pre-training tasks (one for each switch on the wall in Figure \ref{fig:env-screenshots}).
In the bottle domain, we have 2 pre-training tasks (one for each bottle inside the shelf in Figure \ref{fig:env-screenshots}).
The pre-training reward in the switch domain is 0 upon success, and $\exp{(-\|\text{end effector pos.} - \text{target switch pos.}\| - 0.2)} - 1$ otherwise.
The pre-training reward in the bottle domain is $-1 + 0.5 \cdot \mathbbm{1}[\text{door opened}] + 0.5 \cdot \mathbbm{1}[\text{bottle reached}]$.

\subsubsection{Episode Termination and User Feedback}

During the pre-training phase, episodes only end if the task is successfully completed or the environment times out -- we set the terminal flag to true only when the task is successfully completed.
During the online learning phase, episodes also end if the wrong task is performed (e.g., the wrong switch is flipped).

If the episode ends in a success or in completing the wrong task, the user's binary feedback (provided through a button press) is treated as the reward signal.
However, when the episode ends due to a timeout, we automatically generate a negative feedback signal.
The user's button presses matched the automated feedback in 98\% of episodes in the user study.

\subsubsection{Calibration Videos} \label{calibration-vids}

To generate the 6 videos in the switch domain and 8 videos in the bottle domain that are used to calibrate the non-adaptive baseline interface and our method (see Section \ref{drift-exp}), we execute the pre-trained robot policy $\pi_{\psi,\phi}^{\mathrm{spec}}$. 
In the switch domain, we generate successful videos for 3 different switches, with 2 episodes per switch.
In the bottle domain, we generate successful videos for 2 different bottles and 2 different settings of the door (the door can cover either one of the 2 compartments), with 2 episodes in each of the 4 configurations.

\subsubsection{Simulated User Model Parameters} \label{sim-exp-details}

For the simulated user input in Section \ref{ab-exp}, we set the standard deviation of the Gaussian user input noise to 0.1 in the switch domain, and 0.15 in the bottle domain.

\subsubsection{Valve Rotation Experiment Details} \label{valve-app}

We use a similar pre-training procedure as the one described earlier in Appendix \ref{pretrain-rews} for the switch and bottle domains, except that we train for 8000 epochs, train on 50 human demonstrations, initialize the encoder and critic networks by running AWAC~\cite{nair2020accelerating} on the human demonstrations for 25000 iterations, set the pre-training reward function to $\exp{(-5 \cdot |\text{diff. between current and target angle in radians}|)} - 1$, and only terminate an episode upon reaching 200 timesteps.
Each observation includes the 3D end effector position, 4D end effector orientation, 3D end effector velocity, 3D valve position, 2D representation of the current valve orientation ($\sin$ and $\cos$), 1D velocity of the valve joint, 7D arm joint positions, and 7D arm joint velocities.
To speed up human-in-the-loop learning, our encoders only operate on a subset of the observation features: the 3D end effector position, 2D valve orientation, and 3D valve position.
The task specification $\spec$ is the 2D valve orientation.
Each action consists of 7D joint torques, clipped to $[-0.25, 0.25]$, as in the switch and bottle domains.
During the pre-training phase, the initial valve angle is sampled uniformly at random from $[0, 2\pi)$, and the target angle is sampled uniformly at random from the same interval but excluding points within $\frac{\pi}{32}$ radians of the initial angle.
During the calibration phase, the initial angle is always 0, and 7 videos are shown to the user, where the target angle is sampled uniformly at random from the discrete set $\{\frac{\pi}{4} \cdot k\}_{k=1}^7$.
During the online learning phase, the initial angle is always 0, and the target angle is sampled uniformly at random from $[0, 2\pi)$ but excluding points within $\frac{\pi}{16}$ radians of the initial angle.
The user is only allowed to end an episode when the valve angle has been within $\frac{\pi}{16}$ radians of the target angle for 20 consecutive steps.
The episode automatically times out after 200 steps, but the user can still indicate a successful task completion after a timeout.
Note that the task specification for each successful trajectory is extracted from the final state that was actually reached -- since the user only needs to be within $\frac{\pi}{16}$ radians of the target to succeed, we may extract a specification that is near, but not necessarily identical to, the ground-truth specification.
During the online learning phase, we use a batch size of 256, where each batch consists of 128 examples from the calibration dataset and 128 examples from the online dataset.
The simulated static user input consists of a 2D target position on a circle centered at the valve.
The simulated dynamic user input consists of a similar 2D target position, but dynamically adjusted so that it is always at most $\frac{\pi}{8}$ radians from the current valve angle, and also on the shortest arc from the current state to the target state.
The simulated directional user input are 3D one-hot encodings for the actions \{clockwise, counter-clockwise, remain\} -- the remain input is generated when the current valve angle is within $\frac{\pi}{16}$ radians of the target angle.
All the simulated user inputs have i.i.d. isotropic Gaussian noise $\epsilon \sim \mathcal{N}(0, 0.2I)$ added to them at each timestep.
Following prior work on deep set encoders~\cite{rakelly2019efficient}, we generate latent embeddings using an encoder that first processes individual $(\bs_t, \bx_t)$ pairs, and outputs the mean and variance of an isotropic Gaussian for each pair.
The Gaussian factors for the most recent 10 timesteps are multiplied, and the latent embedding for the current timestep is sampled from this new Gaussian.
During training, we sample the latent embedding from this Gaussian.
During online episodes, we set the latent embedding to be the mean of the Gaussian.
We set the dimensionality of the latent embedding space to $d = 2$.
After a maximum number of 3 failed attempts to complete the current task, we automatically timeout and sample a new task.
At the start of each episode, we add uniform random noise of magnitude 0.1 to the horizontal position of the valve, initialize the arm position with the same noisy procedure described in Appendix \ref{switch-bottle-setup}, and do not add noise to any other state variables (e.g., the distance to the wall).

\subsubsection{Puck Pushing Experiment Details} \label{puck-app}

We pre-train for 10000 epochs without any human demonstrations, and set the pre-training reward function to $1-\beta\sigma(\alpha(\text{change in distance of block position from goal} + .5 \cdot \text{change in distance of tool position from block})$, where $\sigma$ is the sigmoid function.
The puck and goal positions are initialized uniformly at random within a square of half extent (0.25, 0.15), which is the area that the end effector can reliably reach.
The positions are constrained to be at least 0.1 apart.
The puck is a cube with a half extent of 0.05, and is kept on the table by a gravity of 10, with mass of 0.1 and friction of 0.5.
Each episode terminates automatically when the robot pushes the puck within 0.05 distance of the goal, or 200 timesteps expires.
Each observation consists of the 3D tool position, 4D tool orientation, 3D block position, and 7D arm joint angles.
The task specification $\spec$ is the 3D position of the goal.
As in the valve domain, note that the task specification for each successful trajectory is extracted from the final state that was actually reached -- since the user only needs to be within 0.05 distance of the target to succeed, we may extract a specification that is near, but not necessarily identical to, the ground-truth specification.
During the online learning phase, we use a batch size of 256, where each batch consists of 128 examples from the calibration dataset and 128 examples from the online dataset.
The simulated user input is a 7D vector of arm joint torques, which is generated using a pre-trained oracle policy.
The simulated user inputs are smoothed using an exponential moving average with smoothing factor $\alpha=0.99$ -- i.e., we set $\bx_t \leftarrow \frac{1}{1-\alpha}(\alpha \bx_{t-1} + (1-\alpha)\bx_t)$.
We use the same deep set encoder architecture as in the valve domain to represent the input encoder, and operate on the most recent 20 timesteps of observations and user inputs.
We set the dimensionality of the latent embedding space to $d = 4$.
As in the switch and bottle domains, after a maximum number of 5 failed attempts to complete the current task, we automatically timeout and sample a new task.

\subsection{Details of User Study} 

\subsubsection{Experiment Design} \label{exp-design}

\begin{figure*}[p]
    \centering
    \includegraphics[width=\linewidth]{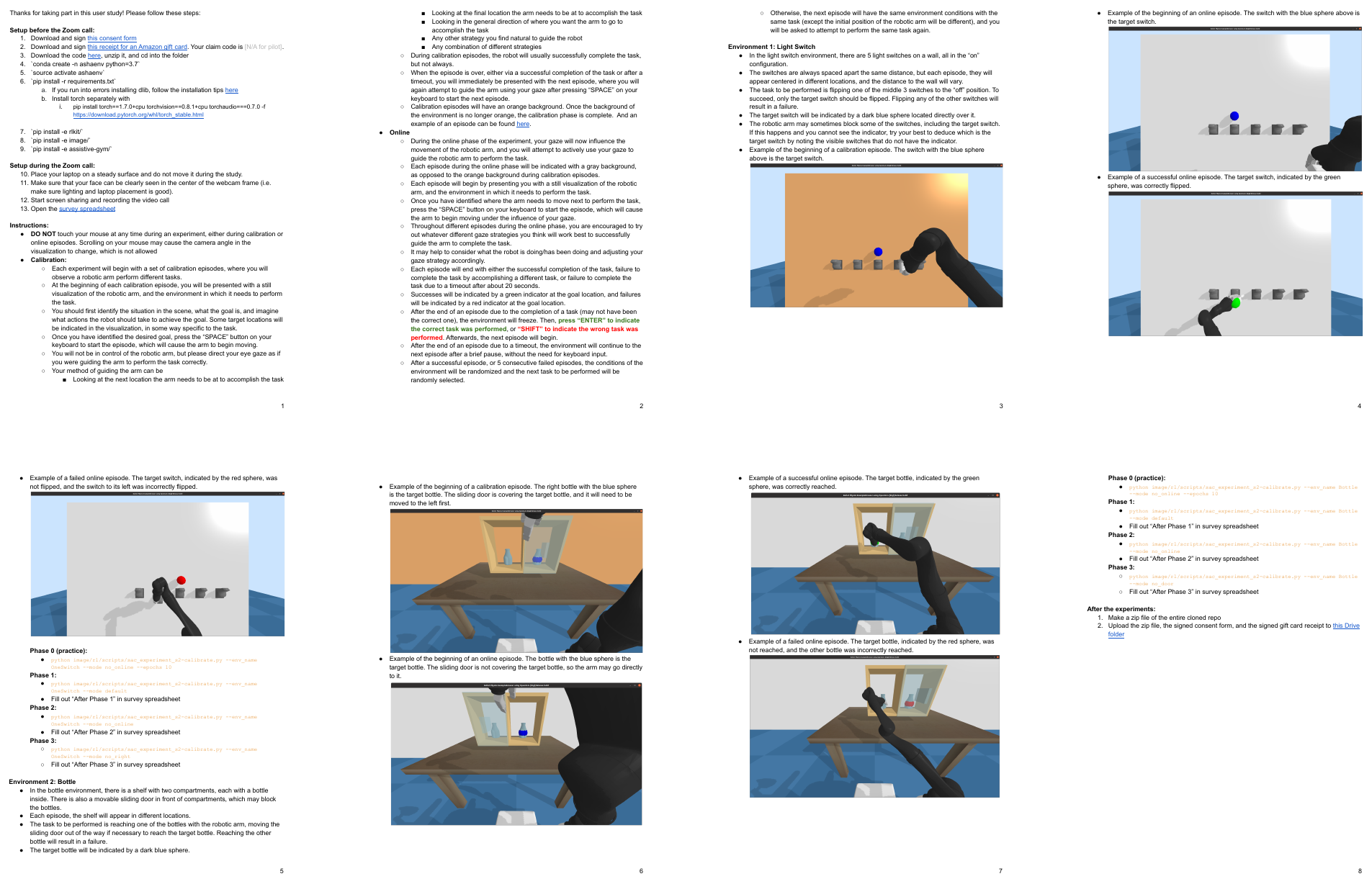}
    \caption{Instructional document for participants in the user study}
    \label{fig:protocol}
\end{figure*}

We recruited 10 male and 2 female participants, with an average age of 21.
Each participant was provided with the rules of each domain (see Figure \ref{fig:protocol}) and a short practice period of 10 episodes to familiarize themselves with the simulation.
Each episode took an average of 13 seconds.
Each participant completed three phases of experiments -- A, B, and C -- in each of the two domains.
Before each phase in the switch domain, each participant generated 6 episodes of calibration data; in the bottle domain, 8 episodes.
In phase A, they use the non-adaptive baseline interface to complete 50 episodes.
In phase B, they use our method to complete 50 episodes.
In phase C in the switch domain, they use our method to complete 50 episodes in which the task distribution is intentionally mismatched with the calibration data (see Section \ref{goal-shift-exp}).
In phase C in the bottle domain, they use our method to complete 50 episodes in which the environment conditions are intentionally mismatched with the calibration data (see Section \ref{env-shift-exp}).
To control for the confounding effects of the user learning or getting fatigued over the course of the full study, we counterbalanced the order of the three phases (i.e., 2 participants followed the order ABC, another 2 participants followed BAC, etc.).

\subsubsection{Subjective Evaluations and Additional Quantitative Results} \label{more-user-study-results}

When prompted to ``please describe your input strategy'', participants responded as follows.

\begin{displayquote}
\small
User 1:\\
Switch Domain:\\
After Phase A: \\
\textbf{gaze at target switch 100\% of the time, hold same gaze throughout}\\
After Phase B: \\
same as phase A\\
After Phase C:\\
same as phase A\\
Bottle Domain:\\
After Phase A:\\
\textbf{If door needs to be opened, then stare at door and then slide gaze to open the door once hand reaches door. After door is opened, stare at vase. If no door needs to be openend, just stare at vase.}\\
After Phase B:\\
same as phase A\\
After Phase C:\\
same as phase A\\
\\
User 2:\\
Switch Domain:\\
After Phase A: \\
\textbf{I looked as far in the left/right direction as possible until the system reached the target}\\
After Phase B: \\
same as phase A\\
After Phase C:\\
same as phase A\\
Bottle Domain:\\
After Phase A:\\
I looked as far in the left/right direction as possible until the system reached the target\\
After Phase B:\\
same as phase A\\
After Phase C:\\
I looked as far in the left/right direction as possible until the system reached the target. \textbf{I stopped looking at the door to ``pull'' it once the system had grasped the handle.}\\
\\
User 3:\\
Switch Domain:\\
After Phase A: \\
\textbf{Look towards a direction I wanted the arm to move in, then stare down the location when I wanted it to flip.}\\
After Phase B: \\
\textbf{exaggerate movement when it was wrong}\\
After Phase C:\\
same as A\\
Bottle Domain:\\
After Phase A:\\
\textbf{continuous movement}\\
After Phase B:\\
\textbf{try to gaze relative to the arm instead of at item}\\
After Phase C:\\
left blank\\
\\
User 4:\\
Switch Domain:\\
After Phase A: \\
\textbf{I looked at the target and if the arm was going in the wrong place I compensated by looking further in the necessary direction}\\
After Phase B: \\
Same strategy as before, \textbf{more compensating because missed more often}\\
After Phase C:\\
Same as before
Bottle Domain:\\
After Phase A:\\
\textbf{Just looked in the direction of the target, sometimes a little bit off to side of the shelf it was on.}\\
After Phase B:\\
Same as before, but more often on target \\
After Phase C:\\
Same as before\\
\\
User 5:\\
Switch Domain:\\
After Phase A: \\
\textbf{look in direction to move}\\
After Phase B: \\
\textbf{exaggerated looking in direction}\\
After Phase C:\\
same\\
Bottle Domain:\\
After Phase A:\\
\textbf{lok left oor right}\\
After Phase B:\\
same\\
After Phase C:\\
\textbf{same, sometimes looko strraright}\\
\\
User 6:\\
Switch Domain:\\
After Phase A: \\
\textbf{Exaggerate look in desired direction and look at target if robot was performing correct task}\\
After Phase B: \\
Same as phase A\\
After Phase C:\\
\textbf{Look at final target}\\
Bottle Domain:\\
After Phase A:\\
\textbf{Exaggerate for left side, look directly at target for right side}\\
After Phase B:\\
Same as phase A\\
After Phase C:\\
Same as phase A\\
\\
User 7:\\
Switch Domain:\\
After Phase A: \\
\textbf{When the robot is close to the goal, gaze at some middle point between robot and goal; when robot is faraway from goal, gaze at some point that goes beyond the goal in the goal direction.}\\
After Phase B: \\
\textbf{Almost always look to the right of the goal; the farther the goal is, the farther the gaze will be from the goal as well.}\\
After Phase C:\\
\textbf{Look at a point slightly beyond the goal in the goal direction and adjust the gaze location according to robot behavior.}\\
Bottle Domain:\\
After Phase A:\\
\textbf{Look at a point beyond the goal in the goal direction, as far as possible, and holding the same gaze; it seems to be working most of the time unless the bottles are far away from the middle.}\\
After Phase B:\\
Same as phase A above.\\
After Phase C:\\
Same as phase A above.\\
\\
User 8:\\
Switch Domain:\\
After Phase A: \\
\textbf{Looked at target and exaggerated/altered gaze if arm moved too far in one direction}\\
After Phase B: \\
Same as above\\
After Phase C:\\
Same as above\\
Bottle Domain:\\
After Phase A:\\
Same as above\\
After Phase B:\\
Same as above\\
After Phase C:\\
left blank\\\\
User 9:\\
Switch Domain:\\
After Phase A: \\
\textbf{I started out looking at the final target, but also tried sweeping my gaze over the intended trajectory and also exaggeratedly looking in the direction I wanted the arm to go}\\
After Phase B: \\
\textbf{I did much the same, but when the arm seemed to be way off the intended trajectory, I'd test out looking in different directions to see how it would affect the trajectory. I wasn't sure how it was actually affecting the trajectory}\\
After Phase C:\\
\textbf{I would try looking directly at the target, then exxagerate my gaze a bit when the arm wasn't doing exactly what I wanted}\\
Bottle Domain:\\
After Phase A:\\
\textbf{I would look at points on the shelf that I wanted the arm to go to, even if not the final bottle. If the bottle was blocked, I'd look at the outer top left corner, then middle of outer edge, then middle divider to move the glass door. If the arm wasn't moving exactly in the direction I wanted, I'd exaggerate my gaze. If it wasn't blocked, I would look directly at the bottle.}\\
After Phase B:\\
\textbf{In this phase, I mainly exaggerated my gaze in the direction I wanted the arm to go, while relying on peripheral vision for tellling where the robot is and relying on the color change for telling if the episode was over}\\
After Phase C:\\
\textbf{I basically had the same strategy as last time, except I exaggerated my gaze even more. I also did the exaggeration in the calibration phase, when I hadn't before. }\\\\
User 10:\\
Switch Domain:\\
After Phase A: \\
left blank\\
After Phase B: \\
left blank\\
After Phase C:\\
left blank\\
Bottle Domain:\\
After Phase A:\\
left blank\\
After Phase B:\\
left blank\\
After Phase C:\\
left blank\\\\
User 11:\\
Switch Domain:\\
After Phase A: \\
\textbf{Gazing in the far extreme direction that I wanted the arm to move towards}\\
After Phase B: \\
\textbf{Gazing at the neighbor of the correct switch in the direction I wanted the arm to move in. (I.e., if I wanted the arm to move farther left, I gazed at the neighbor on the left).}\\
After Phase C:\\
Same as above\\
Bottle Domain:\\
After Phase A:\\
\textbf{Gazing in the extreme direction I wanted the arm to move in}\\
After Phase B:\\
\textbf{I fixed my gaze directly at the bottle I wanted the system to grab.}\\
After Phase C:\\
Same as above\\\\
User 12:\\
Switch Domain:\\
After Phase A: \\
\textbf{I stared at the blue sphere for most of the time and would try to correct the robot if it went to the wrong one by exaggerating my look in a certain direction.}\\
After Phase B: \\
\textbf{I started with the same strategy as in phase 1 where I would stare at the blue sphere above the switch until the arm went there. However, I noticed that this only worked for the switches on the right side, and it wouldn't ever go to the first two switches on the left side--this made me try to compensate for the arm by looking more to the left but it still did not work so I tried different corners of the screen.}\\
After Phase C:\\
\textbf{I started again by trying to stare at the blue sphere and track it with my eyes. However, it seemed like the robot had the opposite problem as phase 2. Instead of always going for the right, it would only go for the second switch on the left. I also tried compensating by looking as far right as I could, but it did not seem to affect the arm.}\\
Bottle Domain:\\
After Phase A:\\
\textbf{If the bottle was behind the glass door, I would stare at the edge of the door and then pretend to slide it with my eyes. Then I would stare at the blue part of the bottle. If the bottle was not behind a door, I looked at the bottle directly.}\\
After Phase B:\\
\textbf{I used a similar strategy to phase 1, where if the bottle was behind the glass door I looked at the edge of the glass door. Once the arm reached the handle of the door, I panned my eyes over to the opposite edge of the box. Once the door was fully open, I would look at the blue part of the bottle.}\\
After Phase C:\\
\textbf{I used my strategy from phase 2 with some slight modifications. I noticed that the arm would get caught on the wrong side of the handle sometimes after opening the door and it would close the door again, so I would sometimes try moving my eyes up and down to get the arm out of the way. There were also times where the arm would open the door half way and then it would move away from the handle, so I tried looking at the opposite corner of the box to keep the arm on the handle.}
\end{displayquote}

These responses show that users employed a variety of different communication styles, including looking directly at the target (users 1, 4, and 8), looking at distant parts of the screen to indicate different targets (users 2 and 3), exaggerating their gaze to correct the robot (users 3-6 and 8), and dynamically guiding the robot to subgoals (users 1, 2, and 7).

\begin{table*}[p]
  \caption{User Study - Subjective Evaluation}
  \centering

\begin{tabular}{lllllll}
\toprule
{} & \multicolumn{3}{c}{Bottle} & \multicolumn{3}{c}{Switch} \\
\cmidrule(lr){2-4} \cmidrule(lr){5-7}
{} &   ASHA & Baseline &     $p$ &        ASHA & Baseline &     $p$ \\
\midrule
\textbf{The system performed the task I wanted} &  4.8 & 3.9 &  $>.1$ &  \textbf{4.2} & \textbf{3.2} &  $\mathbf{<.1}$ \\
I felt in control                                                                                                         &  4.0 & 3.0 &  $>.1$ &  3.6 & 3.1 &  $>.1$ \\
The system responded to my input... &  &  &  &  &  &  \\
...in the way that I expected  &  4.5 & 3.4 &  $>.1$ &  3.5 & 3.2 &  $>.1$ \\
The system was competent at performing tasks... &  &  &  &  &  &  \\
...even if they weren't the tasks I wanted  &  5.2 & 4.9 &  $>.1$ &  5.1 & 4.7 &  $>.1$ \\
\textbf{The system improved over time} &  \textbf{4.9} & \textbf{3.5} &  $\mathbf{<.05}$ &  3.9 & 3.0 &  $>.1$ \\
I improved at using the system over time &  4.0 & 3.2 &  $>.1$ &  3.9 & 3.7 &  $>.1$ \\
I always looked directly at my final target... &  &  &  &  &  &  \\
...holding the same gaze throughout an episode &  4.2 & 3.5 &  $>.1$ &  3.5 & 2.9 &  $>.1$ \\
I compensated for flaws in the system... &  &  &  &  &  &  \\
...by changing my gaze over time &  4.7 & 4.0 &  $>.1$ &  5.4 & 5.5 &  $>.1$ \\
\bottomrule
\end{tabular}
\vspace{5pt}
  \caption*{Subjective evaluations from the 12 participants in the user study. `Baseline' refers to the non-adaptive baseline interface. Means reported below for responses on a 7-point Likert scale, where 1 = Strongly Disagree, 4 = Neither Disagree nor Agree, and 7 = Strongly Agree. $p$-values from a one-way repeated measures ANOVA with the presence of ASHA as a factor influencing responses. While none of the differences shown here are statistically significant, ASHA does outperform the baseline method in terms of the objective metrics analyzed in Section \ref{user-studies}.}
  \label{tab:user-study-survey}
\end{table*}

\begin{figure}[p]
    \centering
    \includegraphics[width=\linewidth]{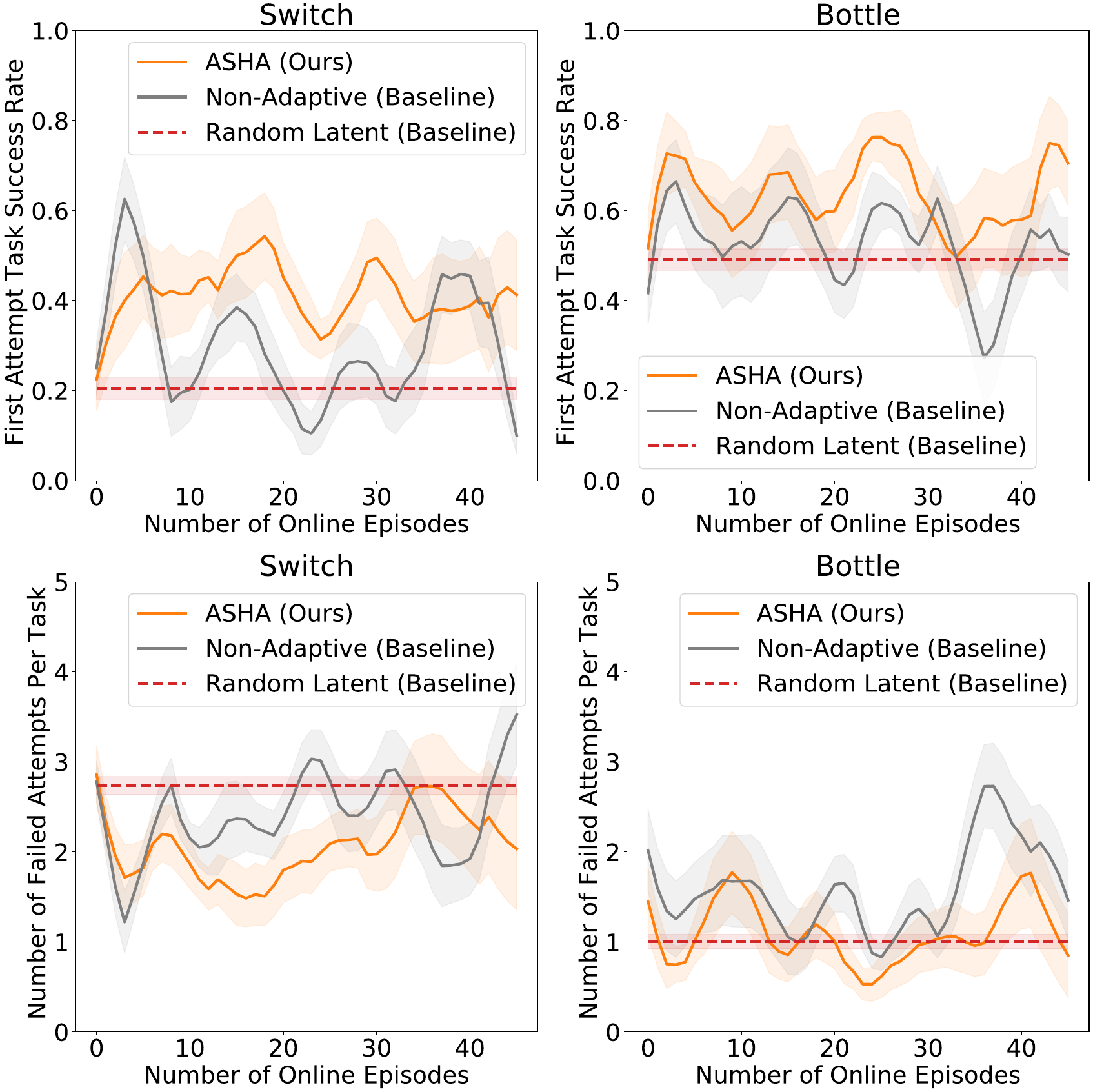}
    \caption{Error bars show standard error across the 12 participants. The maximum number of attempts per task is 5 (see Appendix \ref{timeout}). Both performance metrics -- success rate on the first attempt for each task, and number of failed attempts per task -- generally illustrate similar gaps between ASHA and the baseline methods. However, in the bottle domain, while ASHA achieves a higher success rate than the random-latent baseline, it does not achieve a lower number of failed attempts. This can be attributed to selection effects for difficult tasks in subsequent attempts -- see Figure \ref{fig:user-study-attempts} for details.}
    \label{fig:user-study-line}
\end{figure}

\begin{figure}[p]
    \centering
    \includegraphics[width=\linewidth]{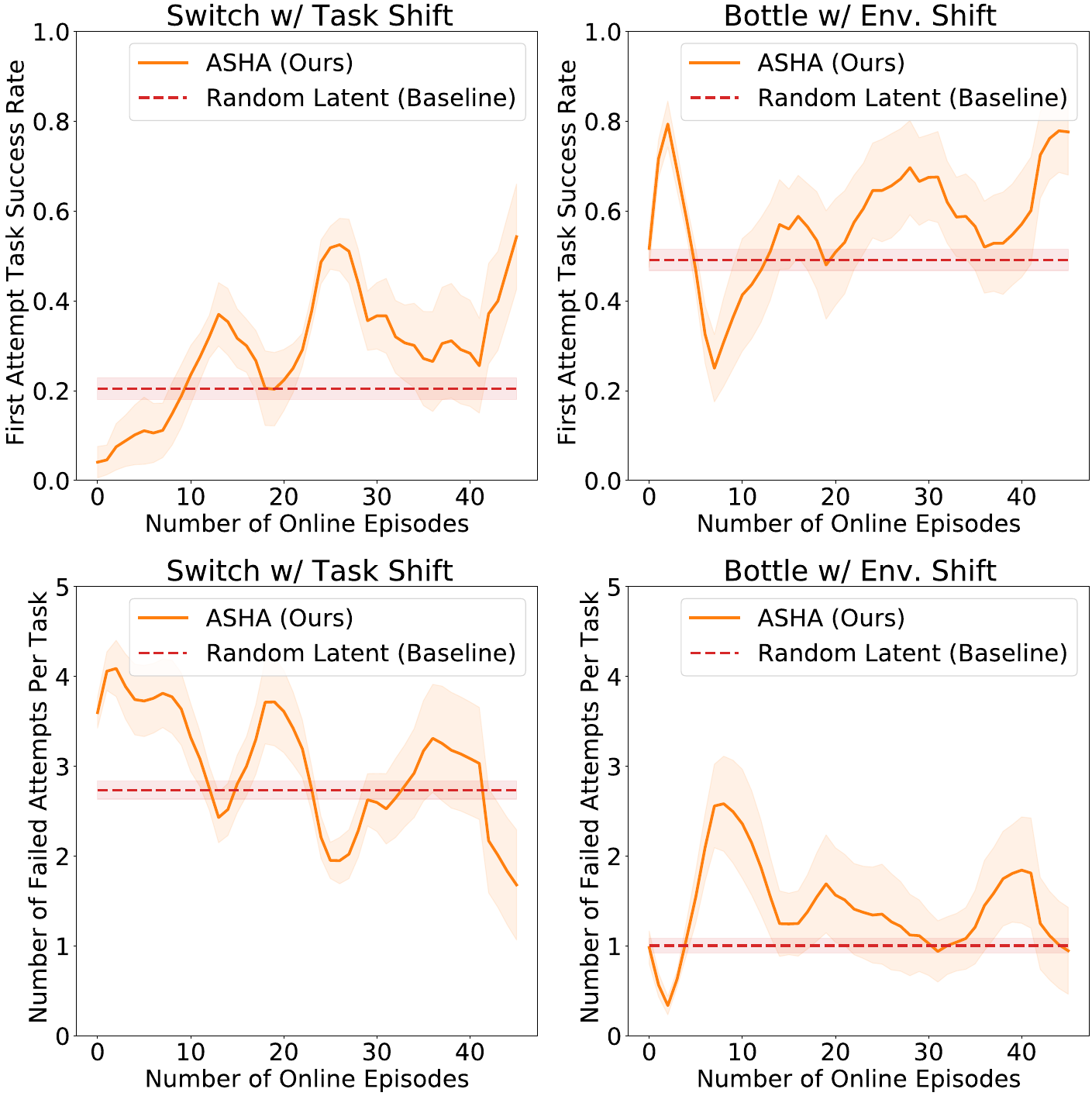}
    \caption{Error bars show standard error across the 12 participants. The maximum number of attempts per task is 5 (see Appendix \ref{timeout}). Both performance metrics -- success rate on the first attempt for each task, and number of failed attempts per task -- generally illustrate similar gaps between ASHA and the random-latent baseline method. However, in the bottle domain, while ASHA achieves a higher success rate than the baseline, it does not achieve a lower number of failed attempts. This can be attributed to selection effects for difficult tasks in subsequent attempts -- see Figure \ref{fig:user-study-attempts} for details.}
    \label{fig:user-study-shift-line}
\end{figure}

\begin{figure}[p]
    \centering
    \includegraphics[width=\linewidth]{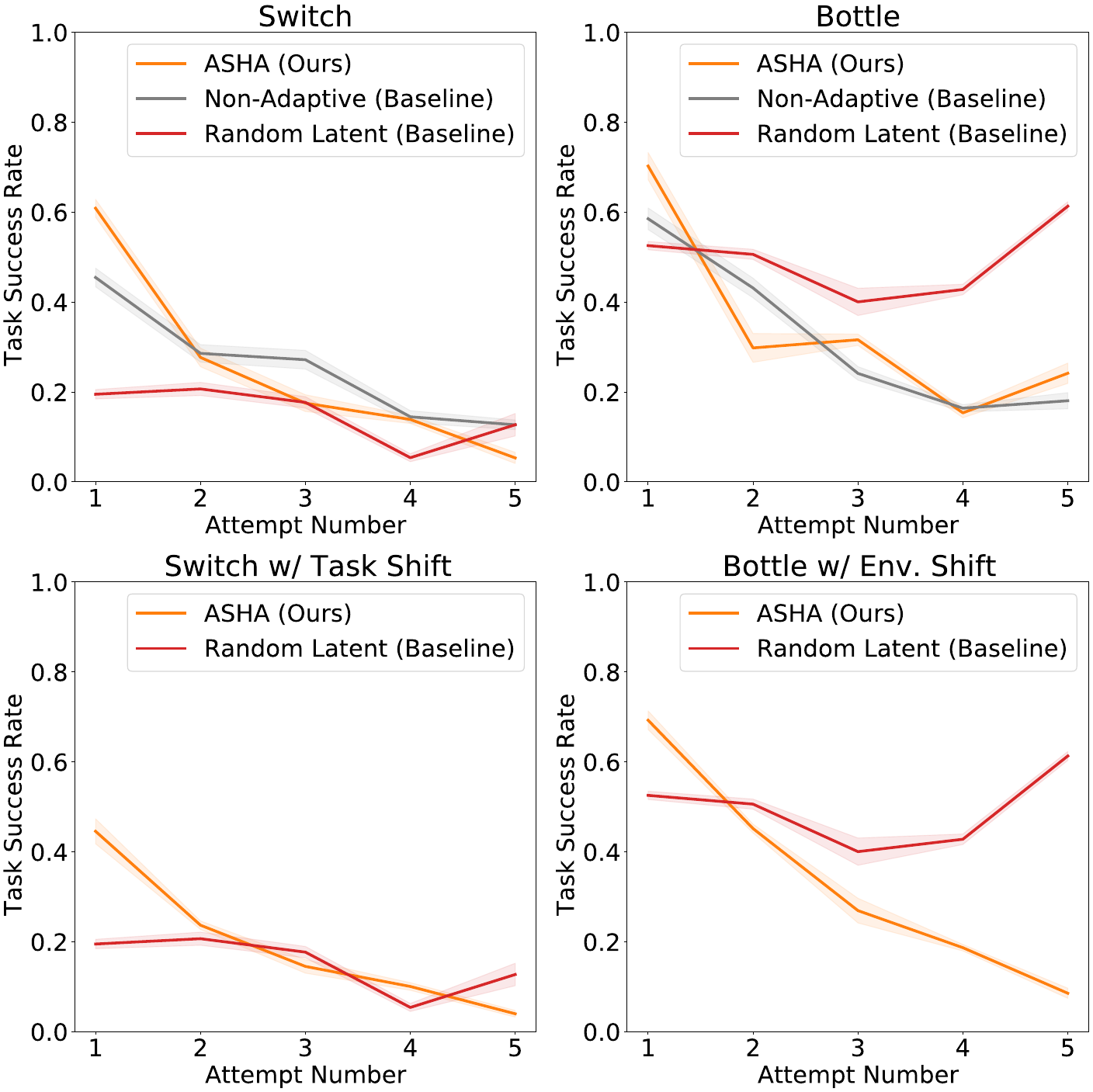}
    \caption{Error bars show standard error across the 12 participants. Performance of ASHA and the non-adaptive baseline tends to decrease on later attempts due to selection effects: tasks in which the user's inputs are easy to interpret for ASHA and the non-adaptive baseline are completed within a small number of attempts, while tasks for which user inputs are difficult to interpret for these two methods tend to require more attempts. Performance of the random-latent baseline is relatively constant across attempts, since it does not take user input, and hence difficult episodes are not selected for in later attempts. On the first attempt, where selection effects do not exist for any of the three methods, ASHA outperforms both the non-adaptive and random-latent baselines.}
    \label{fig:user-study-attempts}
\end{figure}

\begin{figure*}[p]
    \centering
    \includegraphics[width=\linewidth]{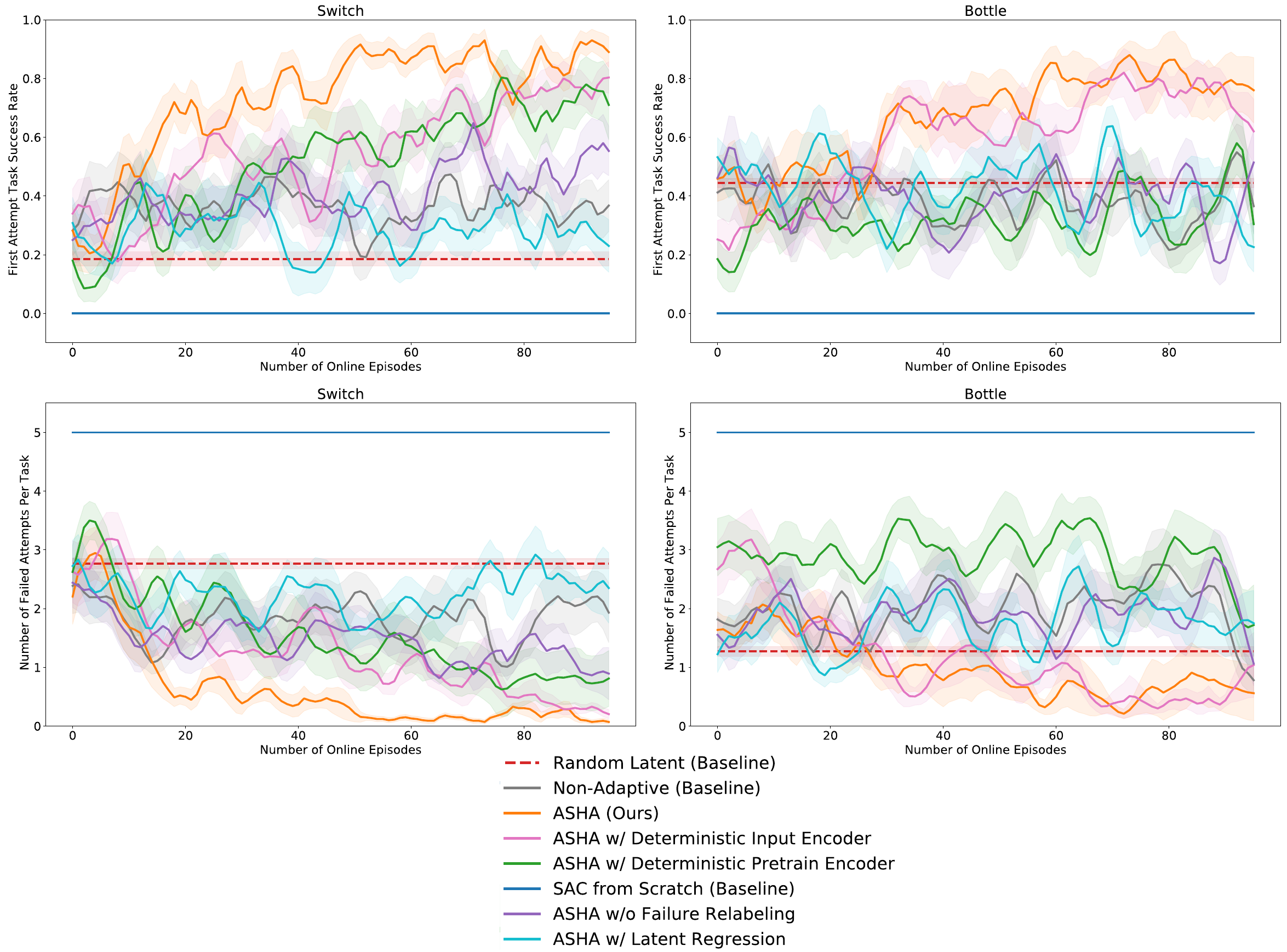}
    \caption{Results from the ablation experiments in Section \ref{ab-exp}. Error bars show standard error across 10 random seeds. The maximum number of attempts per task is 5 (see Appendix \ref{timeout}). As in Table \ref{tab:ablation-means} in Section \ref{ab-exp}, the results show that all the ablated variants of ASHA perform worse than the full ASHA method, suggesting that sampling from a stochastic input encoder $f_{\theta}^{\mathrm{inpt}}$ improves exploration (\textbf{Q1}), pre-training with a VIB and reusing the pre-trained latent-conditioned policy $g_{\phi}$ speed up downstream learning (\textbf{Q2}, \textbf{Q3}), relabeling failures makes human-in-the-loop learning more efficient (\textbf{Q4}), and regressing onto an optimal policy is more effective than regressing onto sampled latents (\textbf{Q5}).}
    \label{fig:ablation-line}
\end{figure*}

\begin{table*}[p]
  \caption{User Study - Quantitative Evaluation}
  \centering
  \begin{tabular}{lllll}
    \toprule
    & \multicolumn{2}{c}{Switch} & \multicolumn{2}{c}{Bottle} \\
    \cmidrule(lr){2-3} \cmidrule(lr){4-5}
     & Success Rate & Failed Attempts & Success Rate & Failed Attempts \\
    \midrule
    Random Latent (Baseline) & $0.20 \pm 0.02$ & $2.7 \pm 0.1$ & $0.49 \pm 0.02$ & $1.0 \pm 0.1$ \\
    Non-Adaptive (Baseline) & $0.41 \pm 0.04$ & $1.8 \pm 0.2$ & $0.65 \pm 0.04$ & $1.8 \pm 0.2$ \\
    ASHA (Ours) & $\mathbf{0.52 \pm 0.04}$ & $\mathbf{1.6 \pm 0.2}$ & $\mathbf{0.74 \pm 0.04}$ & $0.8 \pm 0.2$ \\
    ASHA with Task/Env. Shift (Ours) & $0.43 \pm 0.08$ & $2.1 \pm 0.4$ & $0.74 \pm 0.06$ & $\mathbf{0.6 \pm 0.2}$ \\
    \bottomrule
  \end{tabular}
  \vspace{5pt}
  \caption*{Means measured across 50 episodes, and standard errors measured across the 12 participants. `Failed Attempts' refers to the number of failed attempts per task, for which the maximum value is 5 due to timeouts (see Appendix \ref{timeout}). In the switch domain, `ASHA with Task/Env. Shift' refers to task distribution shift (see Section \ref{goal-shift-exp}). In the bottle domain, `ASHA with Task/Env. Shift' refers to environment shift (see Section \ref{env-shift-exp}). These results show that ASHA outperforms the baselines, not just in terms of learning speed or final performance, but also in terms of cumulative regret throughout the experiment.}
  \label{tab:user-study-means}
\end{table*}

\begin{table*}[p]
  \caption{Ablation Experiments}
  \centering
  \begin{tabular}{lllll}
    \toprule
    & \multicolumn{2}{c}{Switch} & \multicolumn{2}{c}{Bottle} \\
    \cmidrule(lr){2-3} \cmidrule(lr){4-5}
     & Success Rate & Failed Attempts & Success Rate & Failed Attempts \\
    \midrule
    Random Latent (Baseline) & $ 0.19 \pm 0.02 $ & $ 2.8 \pm 0.1 $ & $ 0.44 \pm 0.02 $ & $ 1.3 \pm 0.1 $ \\
    Non-Adaptive (Baseline) & $ 0.50 \pm 0.05 $ & $ 1.3 \pm 0.2 $ & $ 0.53 \pm 0.02 $ & $ 1.3 \pm 0.1 $ \\ \hline
    \textbf{ASHA (Ours)} & $\mathbf{ 0.83 \pm 0.02 }$ & $\mathbf{ 0.3 \pm 0.0 }$ & $\mathbf{ 0.79 \pm 0.03 }$ & $\mathbf{ 0.6 \pm 0.2 }$ \\
    ASHA w/ Det. Input Enc. (\textbf{Q1}) & $ 0.70 \pm 0.03 $ & $ 0.7 \pm 0.1 $ & $ 0.73 \pm 0.02 $ & $ 0.6 \pm 0.1 $ \\
    ASHA w/ Det. Pre-train Enc. (\textbf{Q2}) & $ 0.66 \pm 0.06 $ & $ 1.0 \pm 0.3 $ & $ 0.46 \pm 0.03 $ & $ 2.1 \pm 0.2 $ \\
    SAC from Scratch (\textbf{Q3}) & $ 0.00 \pm 0.00 $ & $ 5.0 \pm 0.0 $ & $ 0.00 \pm 0.00 $ & $ 5.0 \pm 0.0 $ \\
    ASHA w/o Failure Relabeling (\textbf{Q4}) & $ 0.54 \pm 0.03 $ & $ 1.0 \pm 0.1 $ & $ 0.55 \pm 0.02 $ & $ 1.3 \pm 0.1 $ \\
    ASHA w/ Latent Regression (\textbf{Q5}) & $ 0.41 \pm 0.04 $ & $ 1.7 \pm 0.2 $ & $ 0.57 \pm 0.02 $ & $ 1.1 \pm 0.1 $ \\
    \bottomrule
  \end{tabular}
  \vspace{5pt}
  \caption*{Means and standard errors measured across 100 episodes and 10 random seeds. See Figure \ref{fig:ablation-line} in the appendix for more detailed plots. As in Table \ref{tab:ablation-means} in Section \ref{ab-exp}, the results show that all the ablated variants of ASHA perform worse than the full ASHA method, suggesting that sampling from a stochastic input encoder $f_{\theta}^{\mathrm{inpt}}$ improves exploration (\textbf{Q1}), pre-training with a VIB and reusing the pre-trained latent-conditioned policy $g_{\phi}$ speed up downstream learning (\textbf{Q2}, \textbf{Q3}), relabeling failures makes human-in-the-loop learning more efficient (\textbf{Q4}), and regressing onto an optimal policy is more effective than regressing onto sampled latents (\textbf{Q5}).}
  \label{tab:full-ablation-means}
\end{table*}

\end{document}